 \documentclass[final,1p,times]{tempcls}

\usepackage{amsmath,amssymb,amsfonts}
\usepackage{amsopn}
\usepackage{graphicx}
\usepackage{subfigure}
\usepackage{epstopdf}
\usepackage{multirow}
\usepackage{algorithm}
\usepackage{algorithmic}
\usepackage{float}
\usepackage{array}
\usepackage{graphicx}
\usepackage{amsmath,amssymb}

\usepackage{expl3}
\ExplSyntaxOn
\newcommand\latinabbrev[1]{
  \peek_meaning:NTF . {
    #1\@}%
  { \peek_catcode:NTF a {
      #1.\@ }%
    {#1.\@}}}
\ExplSyntaxOff
\def\eg{\latinabbrev{e.g}}
\def\etal{\latinabbrev{et al}}
\def\etc{\latinabbrev{etc}}
\def\ie{\latinabbrev{i.e}}

\journal{}

\begin{document}

\begin{frontmatter}



\title{On The Effect of Hyperedge Weights On Hypergraph Learning}


\author[a,b]{Sheng Huang}

\author[b]{Ahmed Elgammal}
\author[a,c]{Dan Yang\corref{cor1}}
\address[a]{College of Computer Science at Chongqing University, Chonqing, 400044, China}
\address[b]{Department of Computer Science at Rutgers University, Piscataway, NJ, 08854, USA}
\address[c]{School of Software Engineering at Chongqing University Chonqing, 400044, China}

\cortext[cor1]{Corresponding author (Dan Yang): dyang@cqu.edu.cn}

\begin{abstract}
Hypergraph is a powerful representation in several computer vision, machine learning and pattern recognition problems. In the last decade, many researchers have been keen to develop different hypergraph models. In contrast, no much attention has been paid to the design of hyperedge weights. However, many studies on pairwise graphs show that the choice of edge weight can significantly influence the performances of such graph algorithms. We argue that this also applies to hypegraphs. In this paper, we empirically discuss the influence of hyperedge weight on hypegraph learning via proposing three novel hyperedge weights from the perspectives of geometry, multivariate statistical analysis and linear regression. Extensive experiments on ORL, COIL20, JAFFE, Sheffield, Scene15 and Caltech256 databases verify our hypothesis. Similar to graph learning, several representative hyperedge weighting schemes can be concluded by our experimental studies. Moreover, the experiments also demonstrate that the combinations of such weighting schemes and conventional hypergraph models can get very promising classification and clustering performances in comparison with some recent state-of-the-art algorithms.
\end{abstract}


\end{frontmatter}


\section{Introduction}
\vspace{-0.2cm}
As a general version of pairwise graphs, hypergraph learning is commonly used in computer vision, machine learning and pattern recognition areas, \eg~\cite{lsc,sum,phr,higher,supervised,adaptive,nmi,he}, since it represents the similarity relation of data via measuring the similarity between groups of points, which is deemed as a fundamental issue in the aforementioned research areas. Recently, many researchers have been keen to develop different hypergraph models for addressing different tasks, and many impressive hypergraph models were proposed. Hypergraph algorithms can be roughly divided into two categories. The first category uses the spectral clustering techniques to partition the vertices via constructing a simple pairwise graph from the original hypergraph. Representative methods include clique expansion \cite{expansion}, star expansion \cite{expansion} and clique averaging \cite{mean}, \etc. The second category from this category defines a hypergaph Laplacian using analogies from the simple pairwise graph Laplacian. Its representative methods include Zhou's normalized Laplacian \cite{zhou}, Bolla's Laplacian \cite{bolla}, \etc. However, interestingly, as was shown in \cite{hol}, all of the previous algorithms, despite their very different formulations, can be reduced to two graph constructions, the star expansion and the clique expansion, and they are equivalent to each other under specific conditions.
\begin{figure*}[!tbp]
\centering
\subfigure[]{
\centering
\includegraphics[scale=0.18]{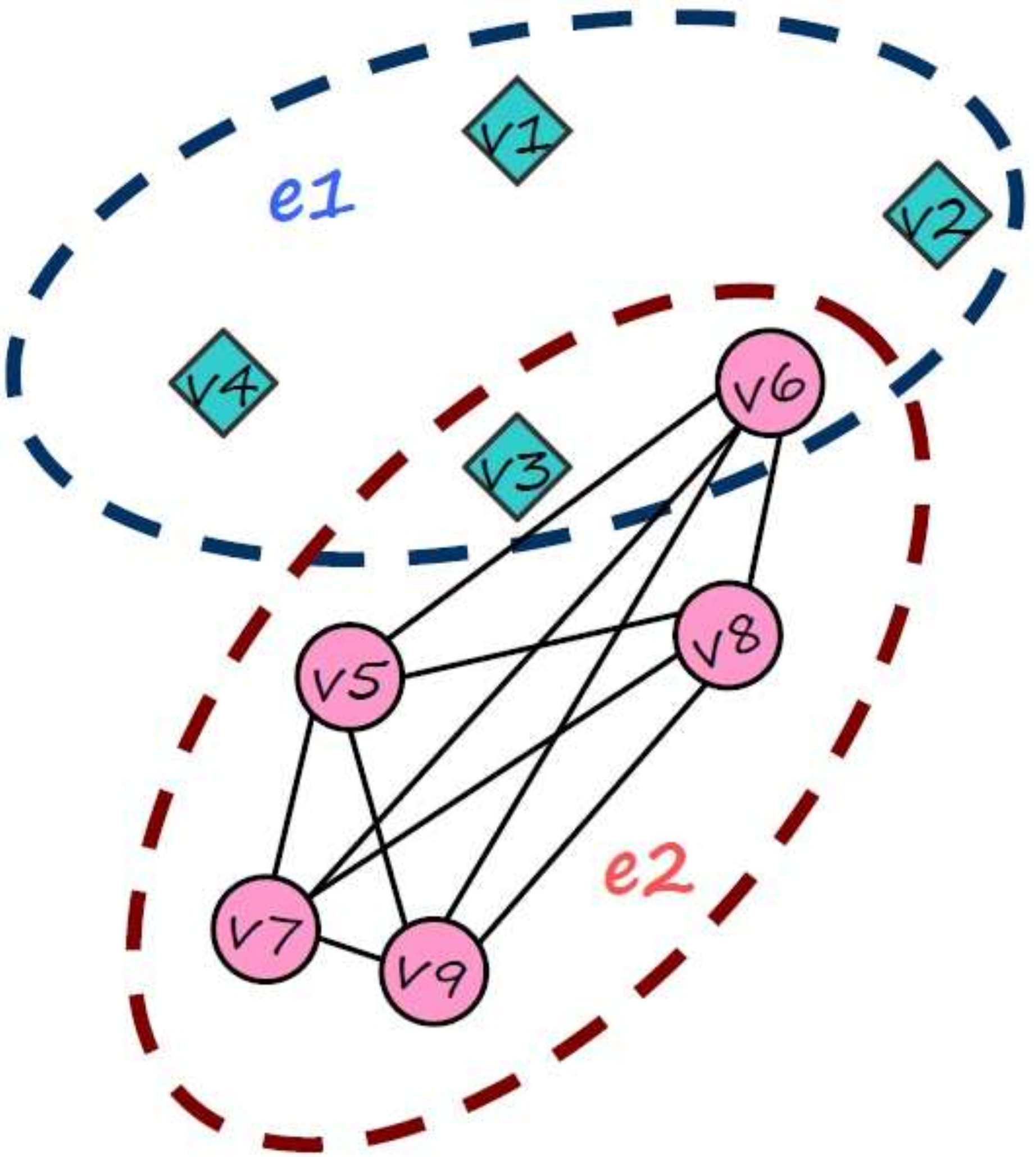}
\label{}}
\centering
\subfigure[]{
\centering
\includegraphics[scale=0.18]{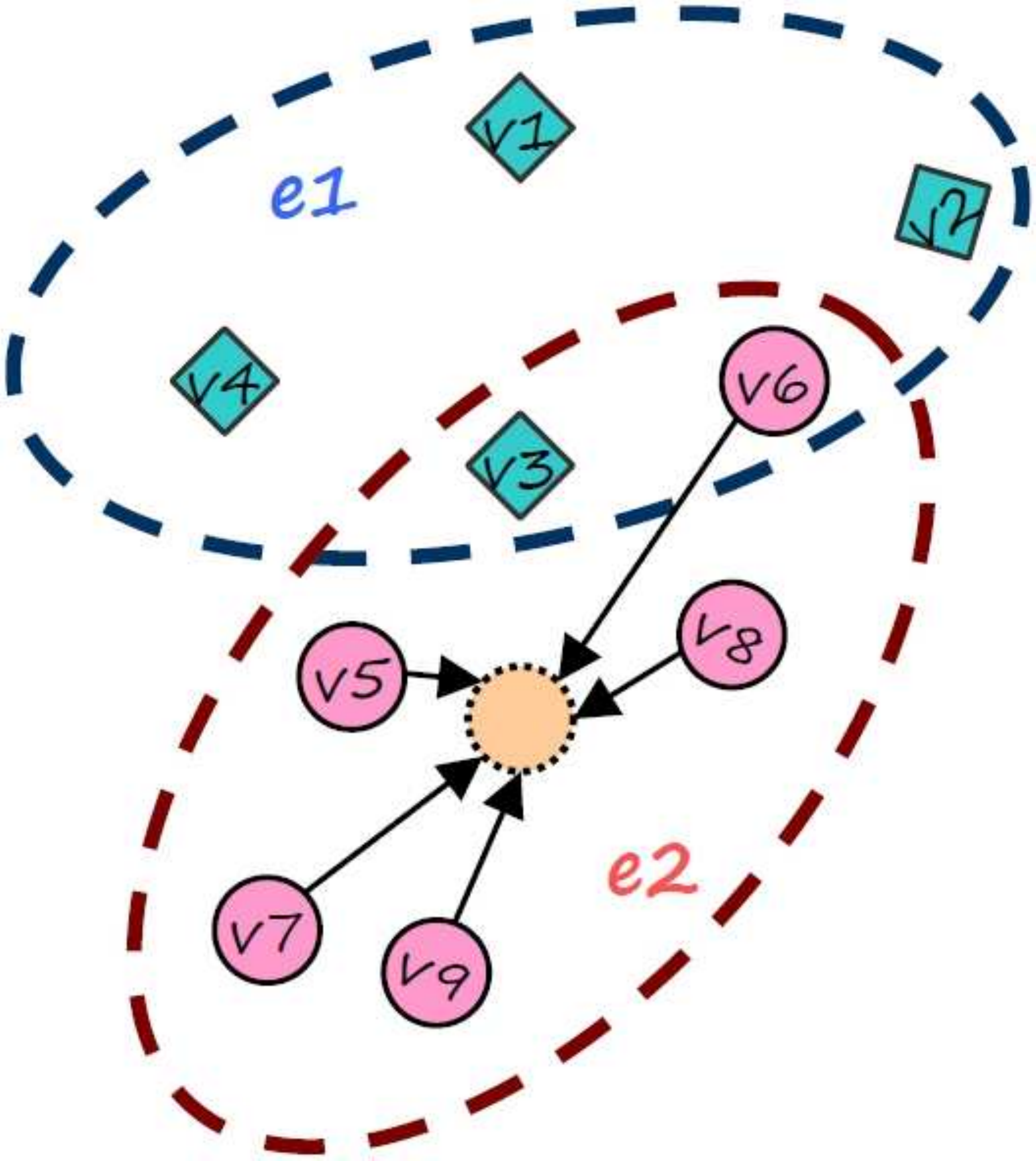}
\label{}}
\centering
\subfigure[]{
\centering
\includegraphics[scale=0.18]{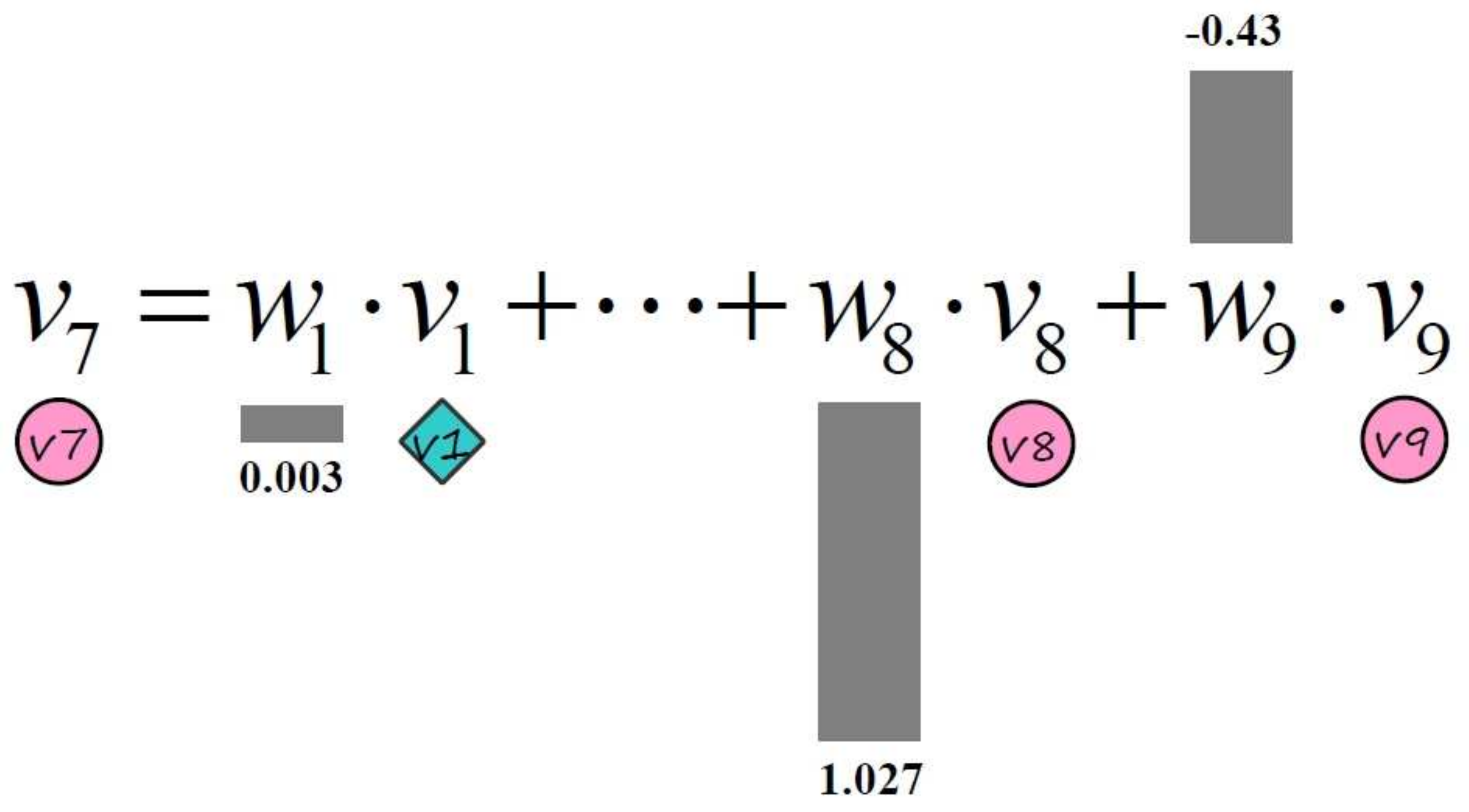}
\label{}}
\vspace{-0.1cm}
\caption{Three explanations of the hyperedge $e2$, (a) a $4$-simplex, (b) a cluster and (c) a linear combination of homogenous vertices.}
\label{example}
\end{figure*}

There are extensive studies about hypergraph construction, \eg~\cite{mean,bolla,rodriguez,zhou,expansion}. But, to the best of our knowledge, there are no prior works to formally discuss the importance of hyperedge weight to hypergraph learning. In the graph learning, which is the pairwise case of hypergraph, extensive studies have already shown that a good choice of edge weight can significantly improve the graph-based algorithms. The Heat-Kernel and Dot-Product weighting schemes are considered as the two most representative weighting schemes of edges \cite{lpi,gnmf,lpp,glpp}. Therefore, we argue that the choice of hyperedge weights also should play a crucial role in hypergraph learning. This motivated us to investigate if there exist a representative hyperedge weighting scheme in hypergraph learning. Moreover, we believe that different hyperedge weights actually provide different ways to explain the hypergraph from different perspectives. In this paper, we try to fill such gap and empirically discuss the influence of hyperedge weight to hypergraph learning via presenting and evaluating three novel hyperedge weighting schemes.

As several hypergraph algorithms have been proposed, a few hyperedge weighting schemes have been heuristically mentioned in such papers. For example, Huang \etal ~\cite{phr} proposed a probabilistic hypergraph-based image retrieval system. In this system, the hyperedge is generated by $k$-nearest neighbour searching and its weight is the sum of the pairwise edge weights between the centroid (seed point) of hyperedge and its neighbours. Zhang \etal ~\cite{nmi} presented an unsupervised hypergraph-based feature selection method, which measures the high-order similarity of the vertices in a hyperedge using multidimensional interaction information (MII). For addressing a 3-D object retrieval task, Gao \etal ~\cite{sum} calculated the hyperedge weight via directly summing the weights of all pairwise edges whose end points are all in the same hyperedge. Clearly, the computation of such hyperedge weight is actually the inverse process of the clique expansion. So, if we use the mean operation to replace the sum operation, such way will be the inverse process of the clique averaging. Different from the previous three methods, Yu \etal ~\cite{adaptive} defined the hyperedge weight as a parameter of the hypergraph model via imposing a sparsity constraint. Thus, the hyperedge weights can be adaptively learned as the graph model optimized. The initial hyperedge weights of this method are constructed by following Huang's way \cite{phr}, and the global optimal weights still cannot be guaranteed. Certainly, there are also some other hyperedge weighting schemes \cite{higher,supervised}, but most of them are associated with very specific tasks.

Complementary to the previously proposed hyperedge weights, we carefully design three novel hyperedge weights from the perspectives of geometry, multivariate statistical analysis and linear regression~\cite{sparse,collabrative} (see Figure \ref{example}). From the perspective of geometry, a hyperedge can be deemed as a high-order simplex \cite{hol}. Thus, the volume of simplex (VOLUME) is a natural hyperedge weight, which provides a reasonable dissimilarity measure for a point set. Motivated by some studies from geometry \cite{det}, we present three ways to compute the volume of simplex for different situations. It is worthwhile to note that these three ways actually define the mathematical relationships between hyperedges and vertices, a hyperedge and its pairwise edges, a hyperedge and its sub-hyperedges, respectively. From the perspective of data mining and multivariate statistical analysis, the hyperedge can be naturally regarded as a cluster in the sample space, thus the trace of the scatter matrix (TRACE) of the samples in the same hyperedge should be a good hyperedge weight. From the perspective of linear regression \cite{sparse,collabrative}, the linear reconstruction error (LLRE) of the homogenous samples should be smaller than the one of the inhomogeneous samples. So, we consider a hyperedge as a small subset of samples, and use the local linear reconstruction error of each point in the hyperedge to measure the similarity of the point set.

In order to verify the importance of hyperedge weighting scheme in hypergraph learning, three state-of-the-art hypergraph models including Zhou's normalized Laplacian \cite{zhou}, clique expansion and star expansion \cite{expansion}, are adopted to evaluate the different hyperedge weights for clustering and classification. Several representative hyperedge weighting schemes for classification and clustering are concluded from our experimental results on ORL, COIL20, Sheffield and JAFFE databases. Such experimental results also demonstrate that a carefully chosen hyperedge weight can significantly improve the performance of hypergraph algorithms. Moreover, we simply apply the combinations of the traditional hypergraph model and the learned representative weighting schemes for image classification and clustering on some larger databases, such as Sence15 and Caltech256 databases. The results show that such simple combination can also get very promising performances in comparison with the state-of-the-art algorithms.

There are mainly three contributions to our work:

\begin{enumerate}
 \setlength{\itemsep}{0pt}
 \setlength{\parskip}{0pt}
 \setlength{\parsep}{0pt}
  \item Three novel weighting schemes, include the volume of simplex (VOLUME), the trace of scatter matrix (TRACE) and the local linear reconstruction error (LLRE), are proposed from three different perspectives, with a desirable property of VOLUME is that it gives the definitions of the relationships between the hyperedge and its vertices, its pairwise edges, and its sub-hypersedges. Extensive experiments show that our proposed weight schemes significantly outperform the conventional weighting schemes in classification and can get a competitive performance in clustering.
 \item We empirically verify the importance of the choice of hyperedge weight on hypergraph learning and draw the researcher's attention to the importance of the design of hyperedge weight.
 \item Representative hyperedge weighting schemes for classification and clustering are experimentally compared and this is very instructive for the hypergraph-based studies.

\end{enumerate}

The rest of paper is organized as follows: Section \ref{s2} presents the background of hypergraph learning; Section \ref{s3} describes the proposed hyperedge weights. Section \ref{s4} shows how to solve classification and clustering tasks using hypegraph; Section \ref{s5} presents the experiments and the conclusion is summarized in Section \ref{s6}.

\begin{figure*}[!tbp]
\centering
\subfigure[]{
\centering
\includegraphics[scale=0.20]{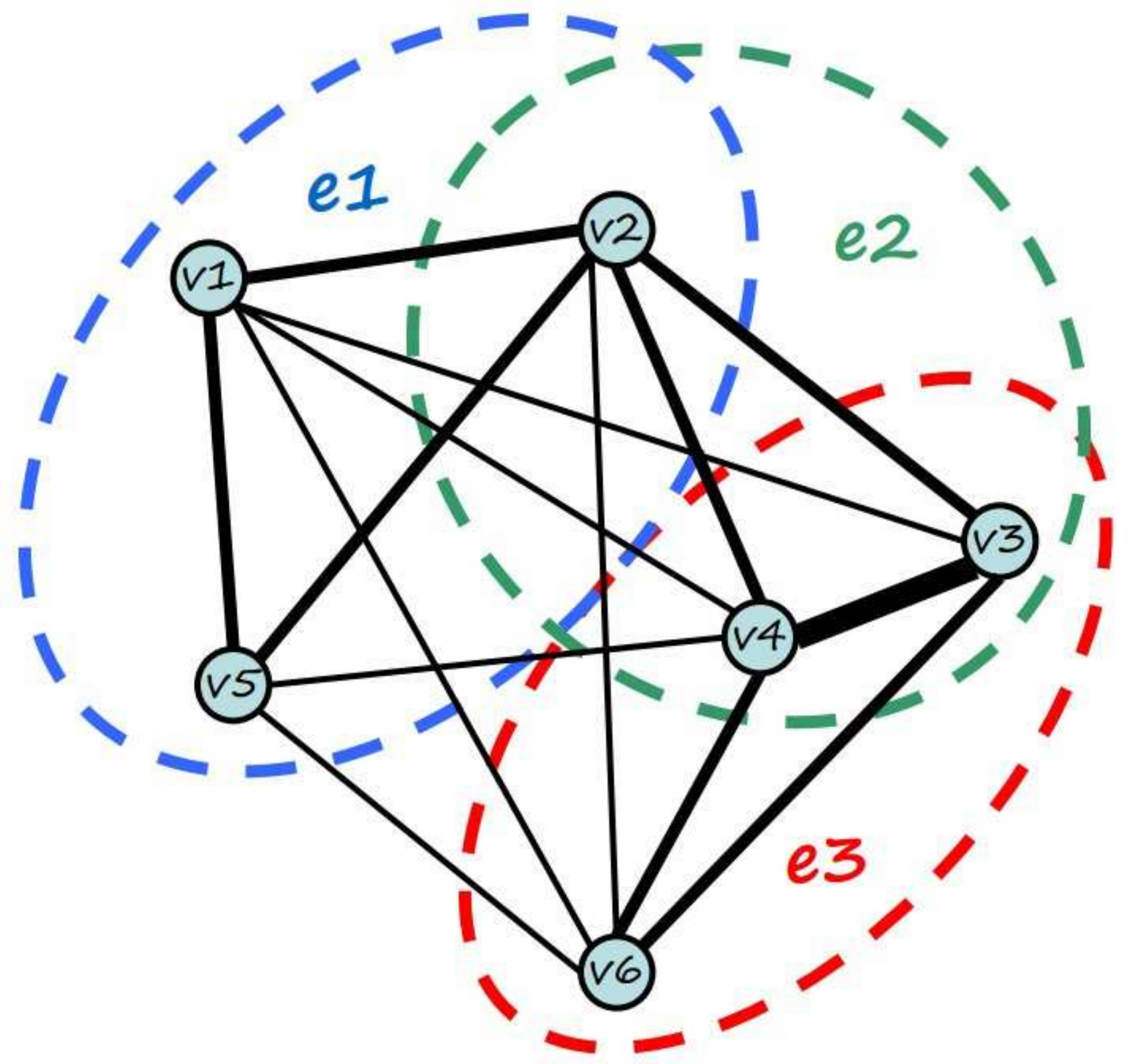}
\label{}}
\subfigure[]{
\centering
\includegraphics[scale=0.28]{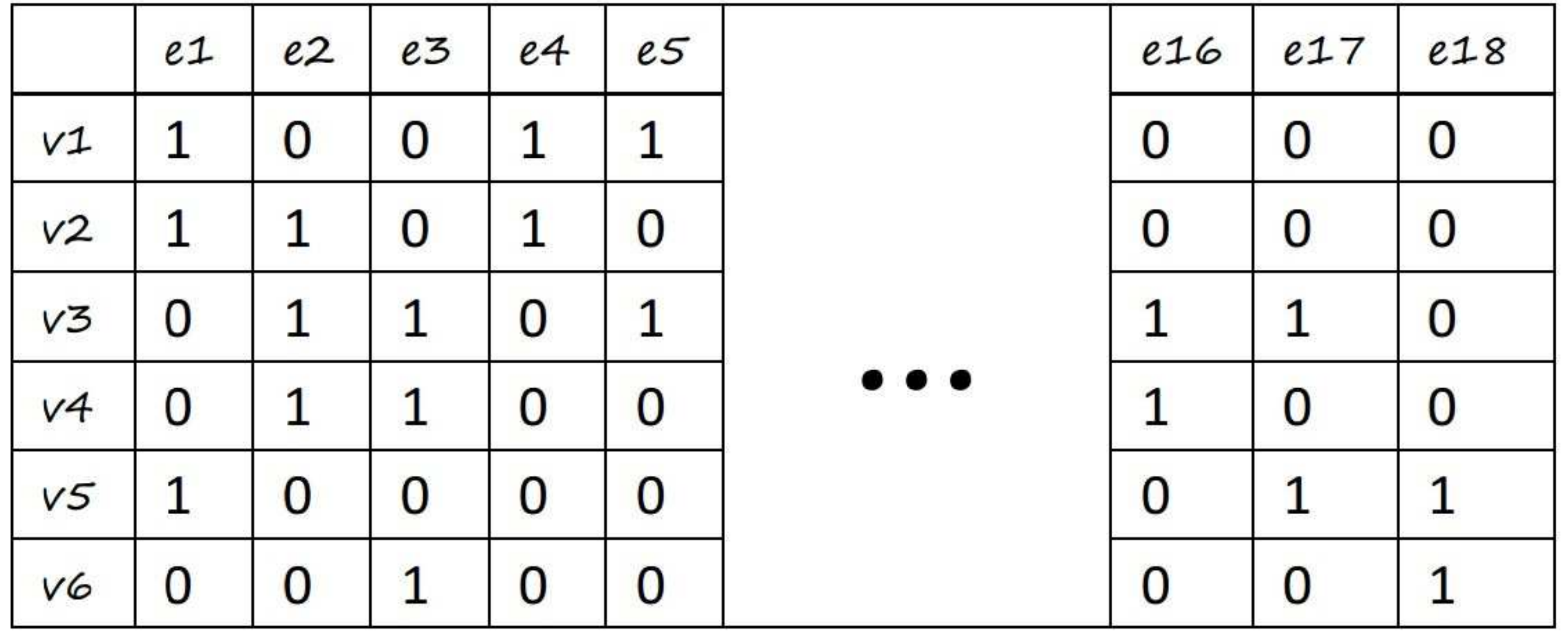}
}
\caption{(a) An example of hypergraph which has 18 hyperedges (15 pairwise edges + 3 three-order hyperedges ), and (b) its corresponding vertex-edge incident matrix.}
\label{hypergraph}
\end{figure*}
\section{Background}\label{s2}
In this section, in order to analyze of the influences of the hyperedge weight to hypergraph learning, the basic notations of hypergraph and three common hypergraph frameworks will be introduced, including Clique Expansion \cite{expansion}, Star Expansion \cite{expansion} and Zhou's normalized hypergraph \cite{zhou}. Besides these three hypergraph frameworks, Bolla's Laplacian \cite{bolla}, Rodriguez's Laplacian \cite{rodriguez} and Clique Averaging \cite{mean} are also very popular hypergraph frameworks. We didn't choose them, since Bolla's Laplacian and Rodriguez's Laplacian are unweighted hypergraphs and Clique Averaging is deemed as solving the same approximation problem as Clique Expansion.
\subsection{Notations}
The key difference between the hypergraph and the ordinary graph is that each edge (hyperedge) of hypergraph can connect more than two vertices (see Figure \ref{hypergraph}). Let $G(V, E)$ denote a hypergraph with vertex set $V$ and edge set $E$. The edges are arbitrary subsets of $V$ with weight $w(e)$ associated with edge $e$. The degree $d(v)$ of a vertex is
\begin{equation}\label{dv}
  d(v) = \sum_{v\in e, e\in E} w(e).
\end{equation}
 The degree of a hyperedge $e$ is denoted by $\delta(e) =|e|$. For $k$-uniform hypergraphs, the degrees of each hyperedge are the same, $\delta(e) =k$. In particular, for the case of ordinary graphs or ¡°2-graphs,¡± $\delta(e) = 2$. The vertex-edge incidence matrix $H$ is $|V|\times|E|$ dimensional binary matrix whose $v,e$-th entry is $h(v,e)$. If $v\in e$, $h(v,e)$ is 1, otherwise it is $0$. By these definitions, we have
 \begin{equation}\label{dv2}
   d(v)=\sum_{v\in e}w(e)h(v,e)
 \end{equation}
 and
 \begin{equation}\label{deltae}
   \delta(e)=\sum_{v\in V}h(v,e).
\end{equation}
$D_e$ and $D_v$ are the diagonal matrices consisting of edge (hyperedge) and vertex degrees, respectively. $W$ is the diagonal matrix of edge weights, $w(\cdot)$.
\subsection{Clique Expansion}
In the clique expansion algorithm \cite{expansion}, each hyperedge is expanded to a clique. A pairwise graph $G_c(V_c,E_c)$ is expanded from the original hypegraph $G(V,E)$ using clique expansion. We have
$V_c = V$ and $E_c = \{(u,v) : u,v\in e,e \in E\}$. The edge weight $w_c(u,v)$ of $G_c$ minimizes the difference between the weight of the graph edge and the weight of each hyperedge $e$ that contains both $u$ and $v$:
\begin{equation}\label{}
w_c(u,v) = {\arg}\underset{w_c(u,v)}{\min}\sum_{u,v\in e, e \in E}(w_c(u,v)-w(e))^2
\end{equation}
The solution of this criterion is
\begin{equation}\label{}
w_c(u,v)=\mu \sum_{u,v\in e, e \in E}w(e)
\end{equation}
where $\mu$ is a fixed scalar. The combinatorial or normalized Laplacian $L_c$ of the constructed graph $G_c$ is then used to partition the vertices.
\subsection{Star Expansion}
In the star expansion algorithm \cite{expansion}, a new vertex is introduced for each hyperedge and this new vertex is connected to each vertex in this hyperedge. More specifically, for a hypergraph $G = (V,E)$, the vertex and edge sets of the star-expanded pairwise-graph, denoted as $V_*$ and $E_*$, are defined as $V_* = V\cup E$ and $E_* =\{(u,e) : u\in e,e \in E\}$, respectively. Thus each hyperedge in
$G$ is expanded into a star in $G_*$, which is a bipartite graph. The weight $w_*(u,e)$ of the edge $(u,e)$ in $G_*$ is given by
\begin{equation}\label{}
  w_*(u,e) = w(e)/\delta(e)
\end{equation}
Since $V_* = V\cup E$, we can assume that in $V_*$, all $v\in V$ are ordered before $e \in E$. Let $M\in R^{|V|\times|E|}$denote the weight between the vertices constructed from $V$ and the vertices from $E$ in $G_*$. The adjacency matrix for $G_*$ can be obtained readily from $M$. Based on the adjacency matrix, the degree for all vertices can be computed. We use $D_{*v}$ and $D_{*e}$ to denote the diagonal matrices of vertex degrees for vertices in $V$ and $E$ in the expanded graph $G_* = (V_*,E_*)$, respectively. Finally, the Laplacian of star expansion is formulated as follows,
\begin{equation}\label{}
L_*=I-D_{*v}^{-1/2}MD_{*e}^{-1}M^TD_{*v}^{-1/2}
\end{equation}
where $I$ is an identity matrix.
\subsection{Zhou's Normalized Laplacian}
Zhou's Normalized Laplacian \cite{zhou} is a representative method of defining the \emph{hypergraph Laplacian} using analogies from the graph Laplacian.  Following the random walk model, Zhou et al. proposed the following normalized hypergraph Laplacian $L_z$:
\begin{equation}\label{}
  L_z=I-D^{-1/2}_vHWD^{-1}_eH^TD^{-1/2}_v
\end{equation}
In the random walk model, given the current position $u\in V$, the walker first chooses a hyperedge $e$ over all hyperedges incident with $u$, with probability proportional to $w(e)$, and then chooses a vertex $v \in e$ uniformly at random.
\section{Hyperedge Weight Computation} \label{s3}
In this section, we design three novel hyperedge weighting schemes from the perspectives of geometry, multivariate statistical analysis and linear regression. The volume of simplex, scatter and linear reconstruction error are adopted respectively as the similarity measure of the point set.

\subsection{Volume of Simplex}
From geometry perspective, each hyperedge can be deemed as a simplex \cite{hol}. Thus, a geometric measure of a set of points can be naturally obtained by computing the volume of the simplex, since a smaller volume of the simplex indicates a closer geometric relationships between the vertices in the hyperedge and vice versa.

There are three ways to compute the volume of the simplex. The first way is to use the vertices of the simplex to compute its volume. Let the vertices of a $k$-degree simplex $E_j$ associated with the $j$th $k+1$-degree hyperedge $e_j$ be represented as $k+1$ $d$-dimensional column vectors $x_0,\cdots,x_{k}$. According to Gram Determinant formula \cite{det}, we cam define a $k\times k$ matrix $G$, whose $i$th column vector $g_i$ is $(x_0-x_i)$. The volume of the simplex can be computed as follows
\begin{equation}\label{}
  Vol(E_j)=\frac{\sqrt{|\det(G^TG)|}}{k!}
\end{equation}
where $\det(\cdot)$ is the matrix determinant and $k!$ is $k$ factorial. This way defines the relationship between the hyperedge weights and its vertices.

The second way is to utilize the edges of the simplex to compute its volumes. This way is very crucial, since it defines the relationship between hyperedge weight and the pairwise edge weights. Let $d_{ij}$ denote the distance between $i$th vertex and $j$th vertex (or to use the pairwise edge weight instead). Then, we can construct $(k+2)\times (k+2)$ presudo-affinity matrix $P$ as follows
\begin{equation}\label{}
P = \left[ {\begin{array}{*{20}{c}}
0&1&1& \cdots &1\\
1&0&{{d_{01}}}& \cdots &{{d_{0k}}}\\
1&{{d_{10}}}&0& \cdots &{{d_{1k}}}\\
 \vdots & \vdots & \vdots & \ddots & \vdots \\
1&{{d_{k0}}}&{{d_{k1}}}& \cdots &0
\end{array}} \right]
\end{equation}
According to the Cayley-Menger Determinant formula \cite{det}, the volume of simplex $E_j$ associated with the hyperedge $e_j$ is denoted as follows
\begin{equation}\label{}
Vol(E_j)=\frac{\sqrt{|\det(P)|}}{2^{k/2} k!}
\end{equation}


The third way is to use the hyperfaces of simplex to compute the volume of simplex. For a $k$-degree simplex, it should have $k+1$ $k$-hyperfaces, where each hyperface is a hyperplane whose Cartesian equation is given by
  \begin{equation}\label{}
a_{i0} + a_{i1} v_1 + a_{i2} v_2 + ... a_{ik} v_k = 0, \quad 0\leq i \leq k
  \end{equation}
where $v_j$ are variables standing for real numbers and the $a_{ij}$ are real constants.
Let $A$ be the $(k+2)\times(k+2)$ matrix with elements $a_{ij}$ and $A_{i0}$ be the cofactor matrix of matrix $A$ with respect to $a_{i0}$.  Then, according to the Klebaner-Sudbury-Satterson Determinant formula \cite{det}, the volume of simplex $E_j$ can be computed as follows
\begin{equation}\label{}
Vol(E_j)=\frac{|\det(A)|^k}{k!\prod_{i=0}^{k}\det(A_{i0})}
\end{equation}

Actually, such cases where only information about hyperfaces is known is very strict, which may seldom happen in practical applications. But we still think this formulation is noteworthy, since it theoretically sets up a link between the sub-hyperedge and hyperedge. The reason why it can put such link is that a hyperface of the simplex is also a simplex, for example, the 2-hyperface of a simplex is a 2-simplex (triangle).

After obtaining the simplex volume, the weight of hyperedge $e_j$ which is associated with the simplex $E_j$ is given as follows
\begin{equation}\label{}
w(e_j)=\exp(-Vol(E_j)/ \mu)
\end{equation}
where $\mu$ is a positive parameter controls the scaling of the hyperedge weight.

The previous formulas held for arbitrary $k$ and $d$. But, the dimensions of feature $d$ should be equal or greater than the degree of hyperedge $k$, \ie,~$k\leq d$ ,since the volumes will be degenerated when $k\ge d$. However, in computer vision applications, typically $k \ll d$ anyway, so the degeneracy is unlikely to happen.

\subsection{Trace of The Scatter Matrix}
From the perspectives of multivariate statistical analysis and data mining, each hyperedge can be considered as a cluster in the sample space. So it is very natural to use the scatter matrix to measure the compactness of a cluster (hyperedge). Therefore, we denote this weight by TRACE. Let the $k\times d$-dimensional matrix $X=[x_1,\cdots,x_k]$ denote the sample matrix associated with the vertices of a $k$-degree hyperedge $e_j$. Then, the scatter matrix $S$, which is a $d\times d$ positive semi-definite matrix, is computed as follows
\begin{equation}\label{}
  S=\sum_{i=1}^k(x_i-\bar{x})(x_i-\bar{x})^T=(X-\bar{X})(X-\bar{X})^T
\end{equation}
where $\bar{x}$ is the mean of the samples and $\bar{X}$ is a $d\times k$-dimensional matrix whose columns are all $\bar{x}$. Finally, we can compute the weight of hyperedge $e_j$ as the trace of the scatter matrix
\begin{equation}\label{}
  w(e_j)=trace(-\exp(S)/\mu)
\end{equation}
where $trace(\cdot)$ denotes the matrix trace, $\exp(\cdot)$ is an element-wise exponential operation and $\mu$ is a positive parameter for controlling the scale of the weight.
\subsection{Local Linear Reconstruction Error}
 We can measure the similarity between a single point and a point set by the linear reconstruction error. The reconstruction error is expected to be smaller if the sample is reconstructed from a homogenous sample rather than an inhomogeneous samples. More specifically, each hyperedge is consider as a subset of the samples. We denote this scheme by LLRE. So, we can follow a leave-one-out strategy to get the reconstruction errors of each sample in such subset via linear regression.
In the case of undirected hypergraph, each vertex will get a reconstruction error. We assume $k$ $d$-dimensional samples $x_1,\cdots,x_k$ are associated with the ordered vertices in $k$-degree hyperedge $e_j$. The reconstruction coefficients $c_i$ of sample $x_i$, which miminizes the reconstruction error can be solved as a least-square problem as follows
\begin{equation}\label{}
\hat{c_i}={\arg}\underset{c_i}{\min}(||x_i-X_{t\neq i,t\in e_j}c_i^T||^2)
\end{equation}
where $X_{t\neq i,t\in e_j}$ is as a $d\times (k-1)$ dimensional matrix whose $t$-th column is $x_t$ where $t \neq i$ and $1\leq t \leq k$. The solution of this problem is $\hat{c_i}=x_i^T(X_{t\neq i,t\in e_j}^T)^\dag$  where $^\dag$ is the generalized inverse of matrix. After obtaining $c_i$, its corresponding reconstruction error $r_i$ can be computed as follows
\begin{equation}\label{}
r_i=\frac{||x_i-X_{t\neq i,t\in e_j}\hat{c_i}^T||^2}{||x_i||^2}
\end{equation}
For an undirected hypergraph, the overall reconstruction error $R$ of a hyperedge can be flexibly assigned as the mean of the reconstruction errors of the samples $R=\frac{1}{k}\sum_{i=1}^{k}r_i$, the minimum of the reconstruction errors of the samples $R=\min(r_1,\cdots,r_k)$ or the maximum of the reconstruction errors of the samples $R=\max(r_1,\cdots,r_k)$. In the case of a directed hypergraph, each directed hyperedge gets one reconstruction error, since the subscript of hypergraph is ordered and only the vertex corresponding to the first subscript of hyperedge is considered as reconstructed point. So, in that case, the overall reconstruction error $R$ of hyperedge is directly equal to $r_i$.
When the hyperedge is generated by $k$-nearest neighbor searching, the overall reconstruction error $R$ of hyperedge can be assigned as the reconstruction error of the samples corresponding to the seed point for saving time. Finally, we use a positive $\mu$ to scale the hyperedge weight with reconstruction error $R$.
\begin{equation}\label{}
  w(e_j)=\exp(-R/\mu)
\end{equation}

Some reasonable constraints can be imposed to the coefficient $c_i$ for furtherly optimizing this model. For example, a sparsity constraint may make sense when the degree of hyperedge is extremely high, thus the coefficient $c_i$ can be solved as a sparse representation task \cite{sparse,sgraph} or collaborative representation task \cite{collabrative}.

\section{Clustering and Classification} \label{s4}
After getting the hyperedge weight, the aforementioned three hypergraph learning frameworks (in section \ref{s2}) are utilized to learn the corresponding hypergraph Laplacians, which can be used for clustering or classification tasks. According to Zhou's work \cite{zhou}, the hypergraph-based clustering and embedding problem is formulated as the following standard Normalized cut (Ncut) problem.
\begin{equation}\label{Ncut}
\small
\hat{f}={\arg}\underset{f\in \mathcal{R}^{|V|}}{\min}f^TLf, \; s.t.\; \parallel f\parallel=1, \langle f,\sqrt{d}\rangle=0
\end{equation}
where the $|V|\times |V|$ dimensional matrix $L$ is the learned hypergraph Laplacian. According to Zhou's work, $\sqrt{d}$ is an eigenvector of $L$ corresponding to the smallest eigenvalue that should be equal to zero. Clearly, this problem can be solved as an eigenvalue problem and the solution of $k$-ways partition are the $k$ eigenvectors of $L$ corresponding to the $k$ smallest nonzero eigenvalues.

With regard to the hypergraph-based classification, the $|V|$-dimensional vector $f$ is deemed as a classification function over $V$, which classifies each vertex
$v$ as the sign $f(v)$. On one hand, in order to assign the same labels to vertices which have many incident hyperedges in common, a functional should be defined to minimize  the sum of the changes of a function over the hyperedges of the hypergraph. According to Zhou's work \cite{zhou}, such functional is exactly as $f^TLf$. On the other hand, the initial label assignment should be changed as little as possible. Let $|V|$-dimensional vector $y_i$ be the label function of the $i$-th class, where $y_i(v)=1$ or -1 if the vertex $v$ belonging to $i$-th class or other classes respectively, and 0 if the vertex $v$ is unlabeled. Thus, the hypergraph-based classification can be formulated as the following optimization problem
\begin{equation}\label{}
  \hat{F}={\arg}\underset{F=[f_1,\cdots,f_c]\in \mathcal{R}^{|V|}} {\min} \sum_{i=1}^{c}(f_i^TLf_i+\lambda ||f_i-y_i||^2)
\end{equation}
where $c$ is the class number and matrix $F$ is the collection of vector $f_i$. $\lambda >0$ is the parameter specifying the tradeoff between the two competitive terms. According to Yu's work \cite{adaptive}, the solution of this problem is as follows
\begin{equation}\label{}
 F=\frac{1}{1+\lambda} {\left( \frac{L+\lambda I}{1+\lambda}\right)^{-1}} Y
\end{equation}
where matrix $Y$ is a label matrix whose $i$-th column is $y_i$. After obtaining $F$, the classification of $i$-th sample can be accomplished by assigning it to the $t$-th class that satisfies $t=argmax_j F_{ij}$.

\section{Experiments} \label{s5}
In order to evaluate the influence of the hyperedge weight choice strategy to hypergraph learning, three classical hypegraphs, namely Zhou's Normalized Laplacian \cite{zhou}, Clique Expansion \cite{expansion} and Star Expansion \cite{expansion}, are used to address the clustering and classification tasks on six databases: ORL\cite{orl}, COIL20 \cite{coil20}, Sheffield \cite{umist}, JAFFE \cite{jaffe}, Scene15 \cite{sence15} and Caltech256 \cite{caltech256} databases. In these experiments, our proposed hyperedge weighting schemes, as well as another three commonly adopted hyperedge weighting schemes are applied to the previous hypergraph frameworks.
\subsection{Data Sets and Experimental Configurations}
Six datasets, including ORL, JAFEE, COIL20, Sheffield, Scene15 and Caltech256-2000, are used in our experiments and their details are reported in Table \ref{dataset}. Among them, Caltech256-2000 dataset \cite{adaptive} is a subset of caltech 256 \cite{caltech256}. We use the first four databases to experimentally study the impact of hyperedge weight to the performance of hypergraph learning, since these four datasets possess manifold structures, and hypergraph learning is a manifold learning technique.
\begin{table*}[!tbp]
\footnotesize
    \caption{The involved datasets }
    \label{dataset}
  \begin{center}
    \begin{tabular}{ c |c |c| c| c |c}
    \hline
    {Database Name} & Classes & Total Samples & Feature & Dimension & Manifold\\
    \hline
    ORL \cite{orl}& 40 & 400 & Grayscale& 10304 & Pose\\
    COIL20 \cite{coil20}& 20 & 1440 & Grayscale & 1024 & View \\
    JAFFE \cite{jaffe}& 10 & 213 & Grayscale & 4096 & Expression \\
    Sheffield \cite{umist} & 20 & 564 & Grayscale & 10304 & Pose \\
    Scene15 \cite{sence15} & 15 & 1500 & PiCoDes \cite{picodes} & 2048& unknown \\
    Caltech256-2000 \cite{adaptive} & 20& 2000 & PiCoDes \cite{picodes}& 2048 & unknown\\
    \hline
    \end{tabular}

  \end{center}
\end{table*}

Three very commonly used hyperedge weighting schemes are implemented to compare with our proposed hyperedge weights. The first hyperedge weighting scheme is 0-1 weighting scheme~\cite{zhou,bolla}, which is also commonly adopted in the regular graph case~\cite{gnmf}. The hypergraph with this weighting scheme can be regarded as the unweighted hypegraph, since all hyperedges in this case are equal to 1. We name this weighting scheme \textbf{BINARY} in our experiments. The second hyperedge weight is the sum of the weights of the pairwise edges in it \cite{sum,expansion},
 \begin{equation}\label{sum_equal}
 w(e)=\exp\left(-\frac{1}{\mu}\sum_{\{v,u\}\in e, v<u}w(u,v)\right).
 \end{equation}
This hyperedge weight computation is the inverse process of Clique Expansion and this weight can be deemed as the \emph{perimeter} of the simplex. For the convenience of discussion, we shortly name this weight \textbf{SUM} in the experiment section. Another frequently used hyperedge weight is the mean of the weights of the pairwise edges in it,
\begin{equation}\label{avg_equal}
  w(e)=\exp\left(-\frac{1}{\mu}{k \choose 2}^{-1}\sum_{\{v,u\}\in e, v<u}w(u,v)\right),
\end{equation}
 where $k$ is the vertex degree of the hyperedge \cite{mean}. This hyperedge weight computation is the inverse process of Clique Averaging. However, this case is actually equivalent to \textbf{SUM}. So, we will not adopt it for comparison. The third hyperedge weighting scheme stems from the KNN-based hyperedge generation. In this case, each hyperedge has a seed point, which is also known as the centroid of the neighborhood. The hyperedge weight is the sum of the distances between the centroid and each of its neighbors in a hyperedge,
 \begin{equation}\label{cent_equal}
    w(e)=\exp\left(-\frac{1}{\mu}\sum_{i\neq c}||x_c-x_j||^2\right),
\end{equation}
where $c$ is the vertex subscript of centroid in hyperedge \cite{phr,adaptive}. For convenience, we shortly name it \textbf{CENTROID}. We remind that our proposed hyperedge weights, based on volume of the simplex, trace of the scatter matrix and local linear reconstruction errors, have been respectively renamed as $\textbf{VOLUME}$, $\textbf{TRACE}$ and $\textbf{LLRE}$ in the introduction.

\subsection{Implementation Details}
It is impracticable to enumerate all possible hyperedges. For example, for a 400 vertices undirected hypergraph, there are more than eight billion 5-degree hyperedges. Therefore, in this paper, we generate hyperedges following Huang's strategy that each hyperedge is generated by a KNN searching given a vertex \cite{phr}. For different databases, the $k$ is different. We set $k=5$ on ORL database and set $k=3$ on COIL20 database following the choice of $k$ in \cite{adaptive}. With regard to Sheffield database and JAFFE database, we apply two-fold cross validation to learn the optimal $k$ which was found to be equal to 5. Similarly, we apply two-fold cross validation to learn the optimal scaling parameter $\mu$ under the different hypergraph frameworks, and we fixed the classification trade off parameter $\lambda$ to 1. With regard to the two larger datasets, Scene15 and Caltech256, we follow the same experimental setting of \cite{adaptive}, where $k=[3,5,10,15,20]$ and $k=[10,20,30,40,50]$ in Caltech256 and Scene15 datasets respectively.
\begin{figure*}[h]
\centering
\subfigure[Zhou's Normalized Hypergraph]{
\centering
\includegraphics[scale=0.19]{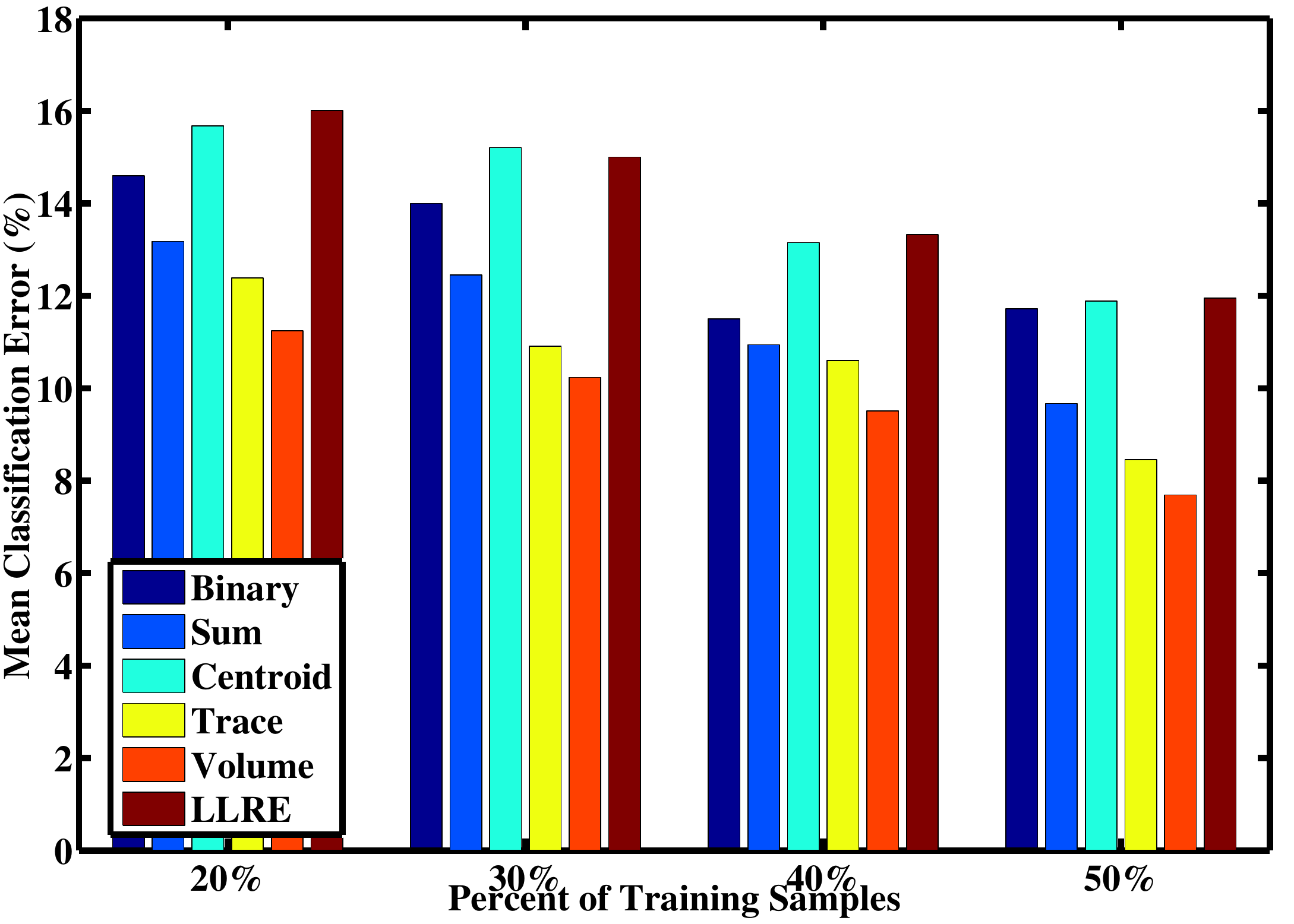}
\label{}}
\subfigure[Clique Expansion]{
\centering
\includegraphics[scale=0.19]{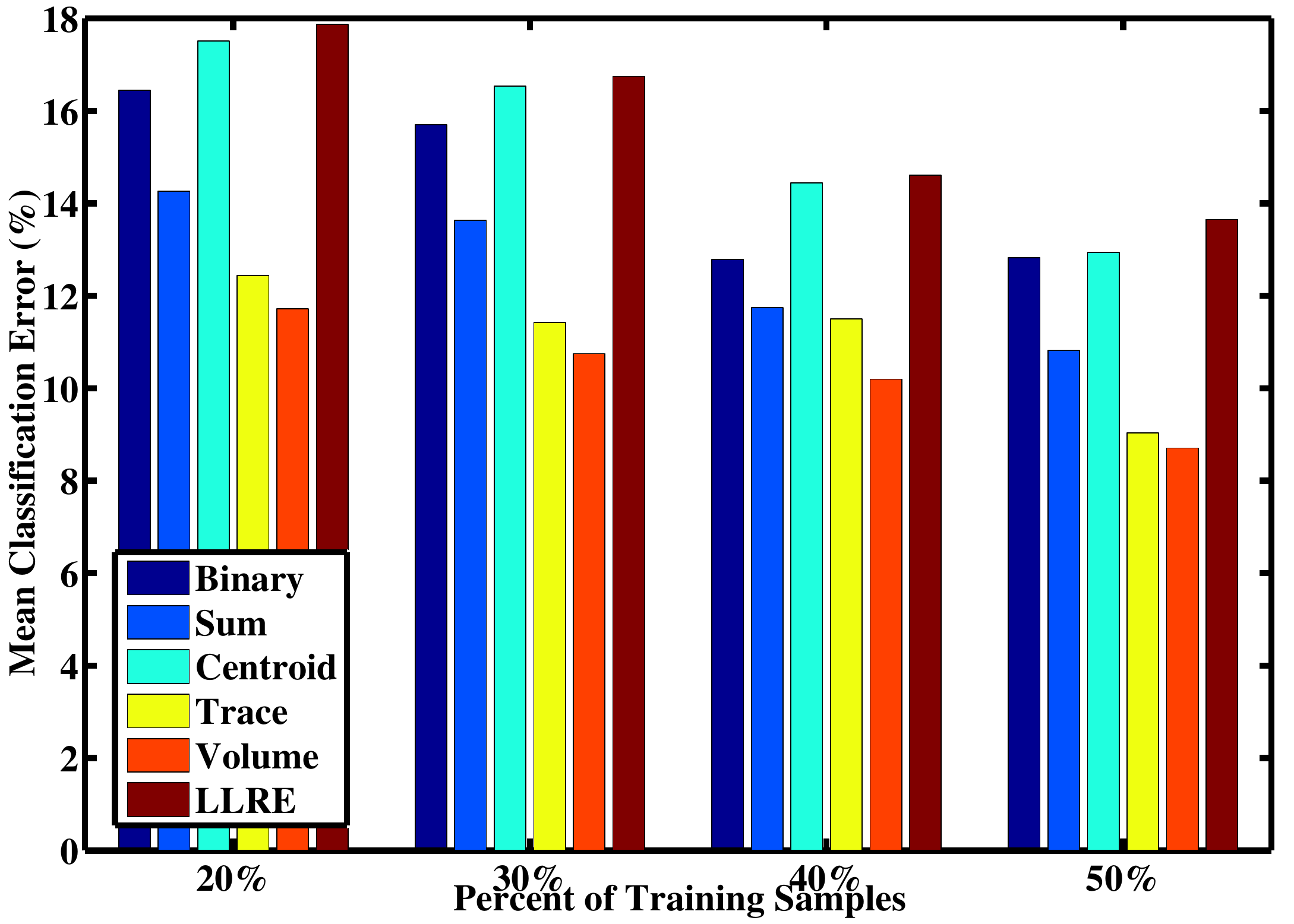}
}
\subfigure[Star Expansion]{
\centering
\includegraphics[scale=0.19]{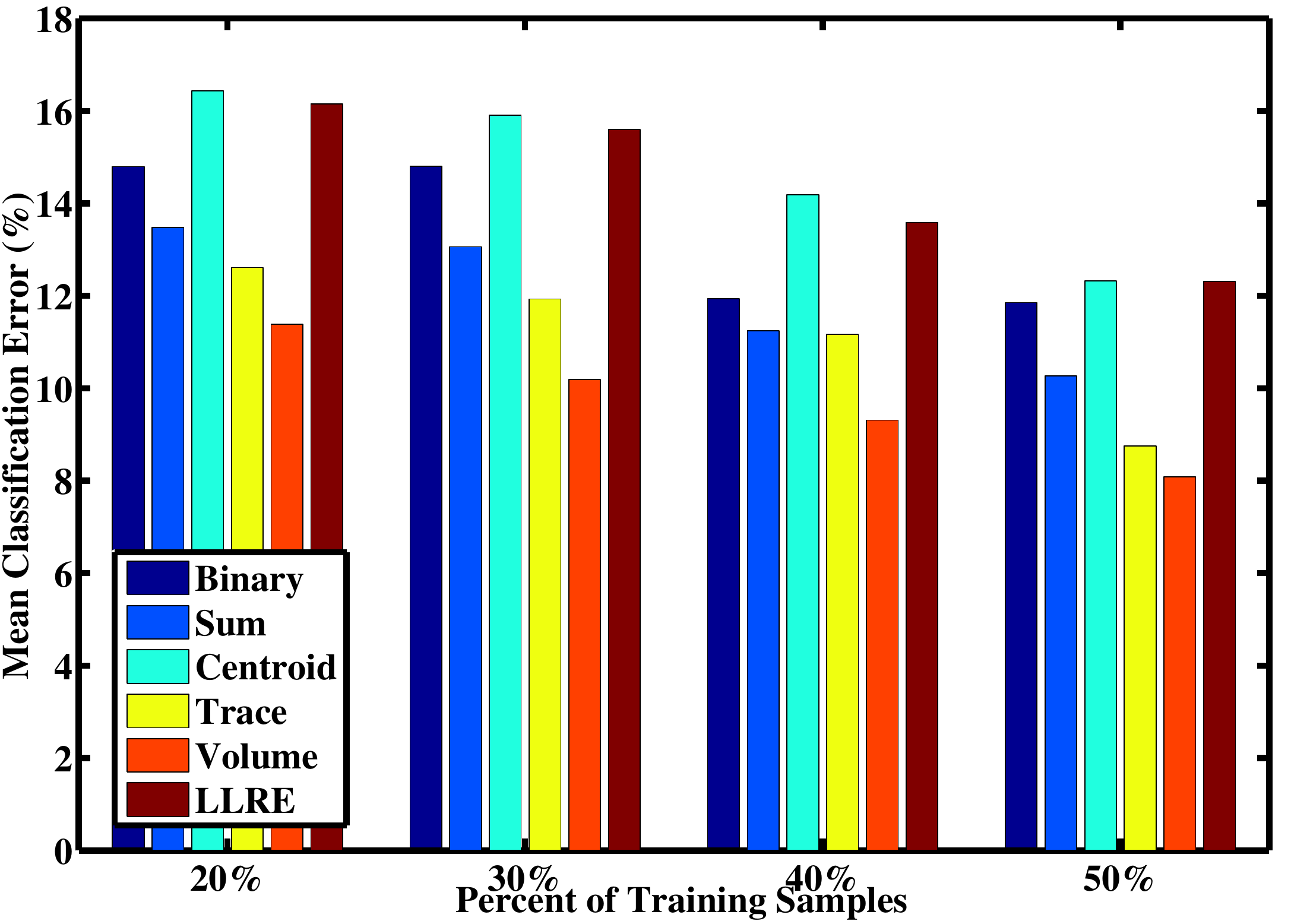}
}
\caption{The mean classification errors of four databases using six different hyperedge weighting schemes respectively under three different hypergraph frameworks. (The lower is better)}
\label{overall_C}
\end{figure*}

\begin{table*}[h]
\footnotesize
  \begin{center}
      \caption{Classification performances of Zhou's Normalized Laplacian using different hyperedge weights on ORL, COIL20, JAFFE and Sheffield databases. (The lower is better) }
    \begin{tabular}{c| p{1.3cm}<{\centering} p{1.3cm}<{\centering} p{1.3cm}<{\centering} p{1.3cm}<{\centering} p{1.3cm}<{\centering} p{1.3cm}<{\centering}}
    \hline
    \multirow{2}*{Database}& \multicolumn{6}{c}{Mean Classification Errors $\pm$ Standard deviation (\%)}\\ \cline{2-7}
    &BINARY & SUM & CENTROID & VOLUME &TRACE&LLRE\\
    \cline{1-7}
ORL&7.75$\pm$1.77&6.25$\pm$1.06&7.75$\pm$1.77&5.50$\pm$1.41&\textbf{4.50$\pm$1.41}&7.25$\pm$1.77\\
JAFEE&15.00$\pm$3.93&\textbf{12.22$\pm$1.57}&15.00$\pm$2.36&12.78$\pm$0.79&13.33$\pm$0.00&13.89$\pm$2.36\\
Sheffield&18.79$\pm$0.73&15.34$\pm$1.71&17.59$\pm$3.90&\textbf{8.79$\pm$3.66}&11.21$\pm$2.68&18.79$\pm$0.70\\
COIL20&5.35$\pm$0.88&4.86$\pm$0.00&7.22$\pm$2.55&\textbf{3.68$\pm$0.69}&4.79$\pm$0.29&7.85$\pm$4.03\\
\hline
Average&11.72&9.67&11.89&\textbf{7.69}& 8.46&11.94\\
    \hline
    \end{tabular}

    \label{C_Zhou}
  \end{center}
\end{table*}

\begin{table*}[h]
\footnotesize
    \caption{Classification performances of Clique Expansion using different hyperedge weights on ORL, COIL20, JAFFE and Sheffield databases.(The lower is better) }
  \begin{center}
    \begin{tabular}{c| p{1.3cm}<{\centering}  p{1.3cm}<{\centering} p{1.3cm}<{\centering} p{1.3cm}<{\centering} p{1.3cm}<{\centering} p{1.3cm}<{\centering}}
    \hline
    \multirow{2}*{Database}& \multicolumn{6}{c}{Mean Classification Errors $\pm$ Standard deviation (\%)}\\ \cline{2-7}
     & BINARY & SUM & CENTROID & VOLUME&TRACE &LLRE\\
    \cline{1-7}
ORL&9.75$\pm$0.35&7.75$\pm$0.35&9.75$\pm$0.35&6.75$\pm$0.35&\textbf{5.75$\pm$0.35}&9.75$\pm$0.35\\
JAFEE&16.11$\pm$5.50&\textbf{14.44$\pm$1.57}&15.00$\pm$3.93&15.00$\pm$0.79&\textbf{14.44$\pm$1.57}&16.11$\pm$5.5\\
Sheffield&20.00$\pm$1.95&15.52$\pm$2.44&18.97$\pm$2.44&\textbf{8.97$\pm$3.90}&11.21$\pm$2.19&20.00$\pm$1.95\\
COIL20&5.42$\pm$0.98&5.56$\pm$0.00&8.06$\pm$2.16&\textbf{4.10$\pm$0.29}&4.72$\pm$0.59&8.75$\pm$3.34\\
\hline
Average&12.82&10.82&12.94&\textbf{8.70}&9.03&13.65\\
    \hline
    \end{tabular}

    \label{C_Clique}
  \end{center}
\end{table*}

\begin{table*}[h]
    \caption{Classification performances of Star Expansion using different hyperedge weights on ORL, COIL20, JAFFE and Sheffield databases. (The lower is better) }
\footnotesize
  \begin{center}
    \begin{tabular}{c| p{1.3cm}<{\centering} p{1.3cm}<{\centering} p{1.3cm}<{\centering} p{1.3cm}<{\centering} p{1.3cm}<{\centering} p{1.3cm}<{\centering} }
    \hline
    \multirow{2}*{Database}& \multicolumn{6}{c}{Mean Classification Errors $\pm$ Standard deviation (\%)}\\ \cline{2-7}
     &BINARY& SUM  & CENTROID & VOLUME &TRACE&LLRE\\
    \cline{1-7}
ORL&8.00$\pm$1.41&6.25$\pm$0.35&8.00$\pm$1.41&5.50$\pm$1.41&\textbf{4.75$\pm$1.06}&7.25$\pm$1.77\\
JAFEE&14.44$\pm$3.14&13.33$\pm$0.00&14.44$\pm$3.14&\textbf{12.78$\pm$0.79}&13.89$\pm$0.79&14.44$\pm$3.14\\
Sheffield&19.48$\pm$0.24&  16.21$\pm$1.46&19.48$\pm$0.24&\textbf{10.17$\pm$3.66}&11.38$\pm$2.93&19.48$\pm$0.24\\
COIL20&5.49$\pm$1.08& 5.28$\pm$0.59&7.36$\pm$2.75&\textbf{3.89$\pm$0.98}&5.00$\pm$0.59&8.06$\pm$4.32\\
\hline
Average&11.85&10.27&12.32&\textbf{8.08}&8.75&12.31\\
    \hline
    \end{tabular}
    \label{C_Star}
  \end{center}
\end{table*}

\subsection{Evaluation in Classification}
\label{hyperclass}
We conduct some experiments to study the influence of hyperedge weight to the hypergraph in classification. The cross validation scheme is applied in these experiments.

Tables~\ref{C_Zhou},~\ref{C_Clique} and~\ref{C_Star} respectively report the classification errors of Zhou's Normalized Laplacian, Clique Expansion and Star Expansion frameworks using different hyperedge weights in two-fold cross validation case. Figure~\ref{overall_C} presents the comprehensive evaluation results of the six weighting schemes under three hypergraph frameworks. The Y-axis of this Figure indicates the mean classification errors of four databases. According to Tables~\ref{C_Zhou},~\ref{C_Clique},~\ref{C_Star} and Figure~\ref{overall_C}, it is clear that the proposed weighting schemes, VOLUME and TRACE, outperforms other four weighting schemes. For example, on Sheffield database, the classification accuracy gains of VOLUME over BINARY, SUM, CENTROID are 11.03\%, 6.45\% and 10\% respectively using Clique Expansion. Such gains for the TRACE are 8.79\%, 4.31\% and 7.76\%. From comprehensive perspective, the average classification accuracy gains of VOLUME over the frequently adopted weighting scheme CENTROID are 4.2\%, 4.24\% and 4.24\% using Zhou's Normalized Hypergraph, Clique Expansion and Star Expansion respectively. These numbers of TRACE are 3.43\%, 3.91\% and 3.47\%.

\begin{figure*}[!tbp]
\centering
\subfigure[20\% samples for training]{
\centering
\includegraphics[scale=0.25]{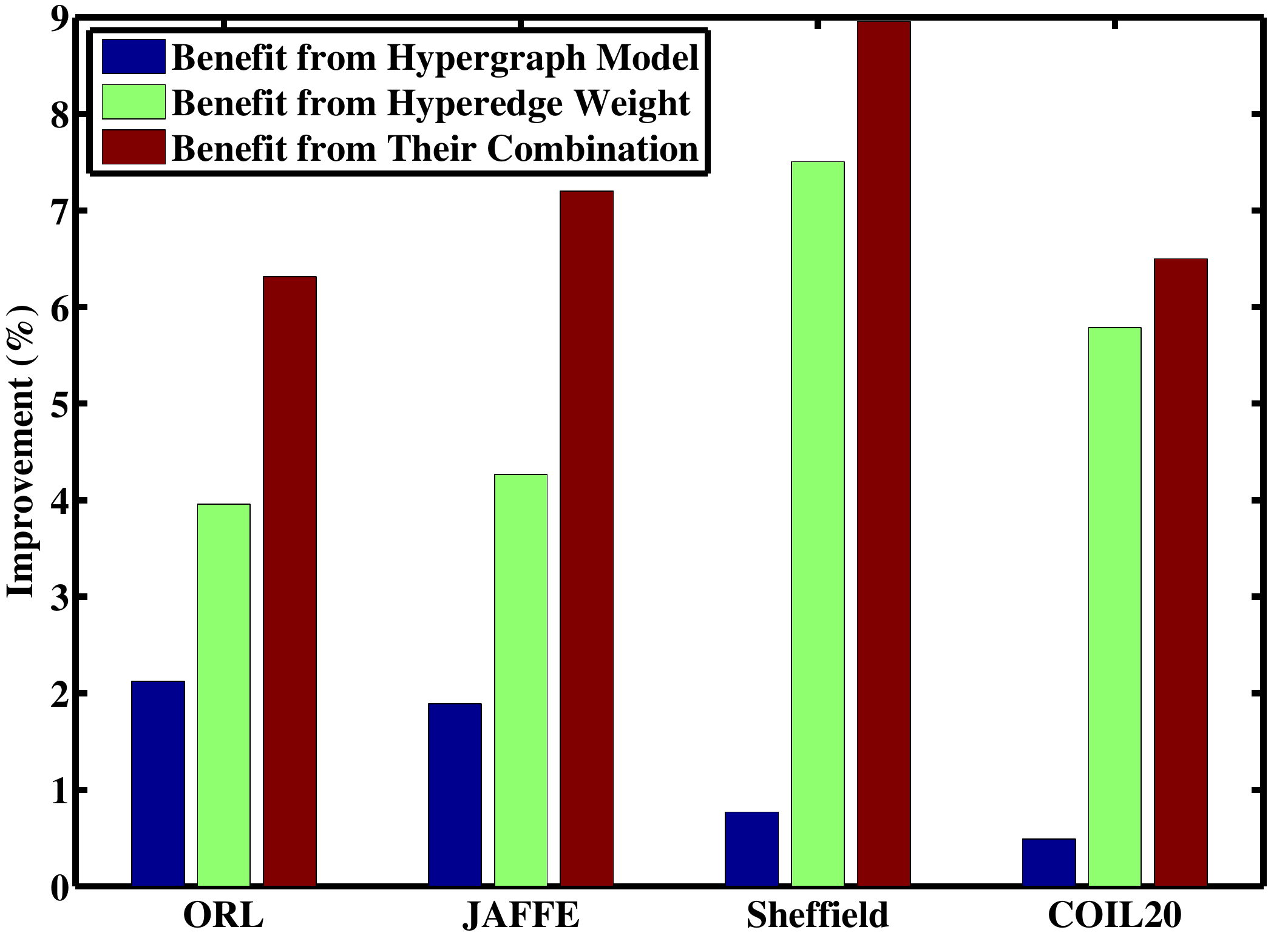}
\label{}}
\subfigure[30\% samples for training]{
\centering
\includegraphics[scale=0.25]{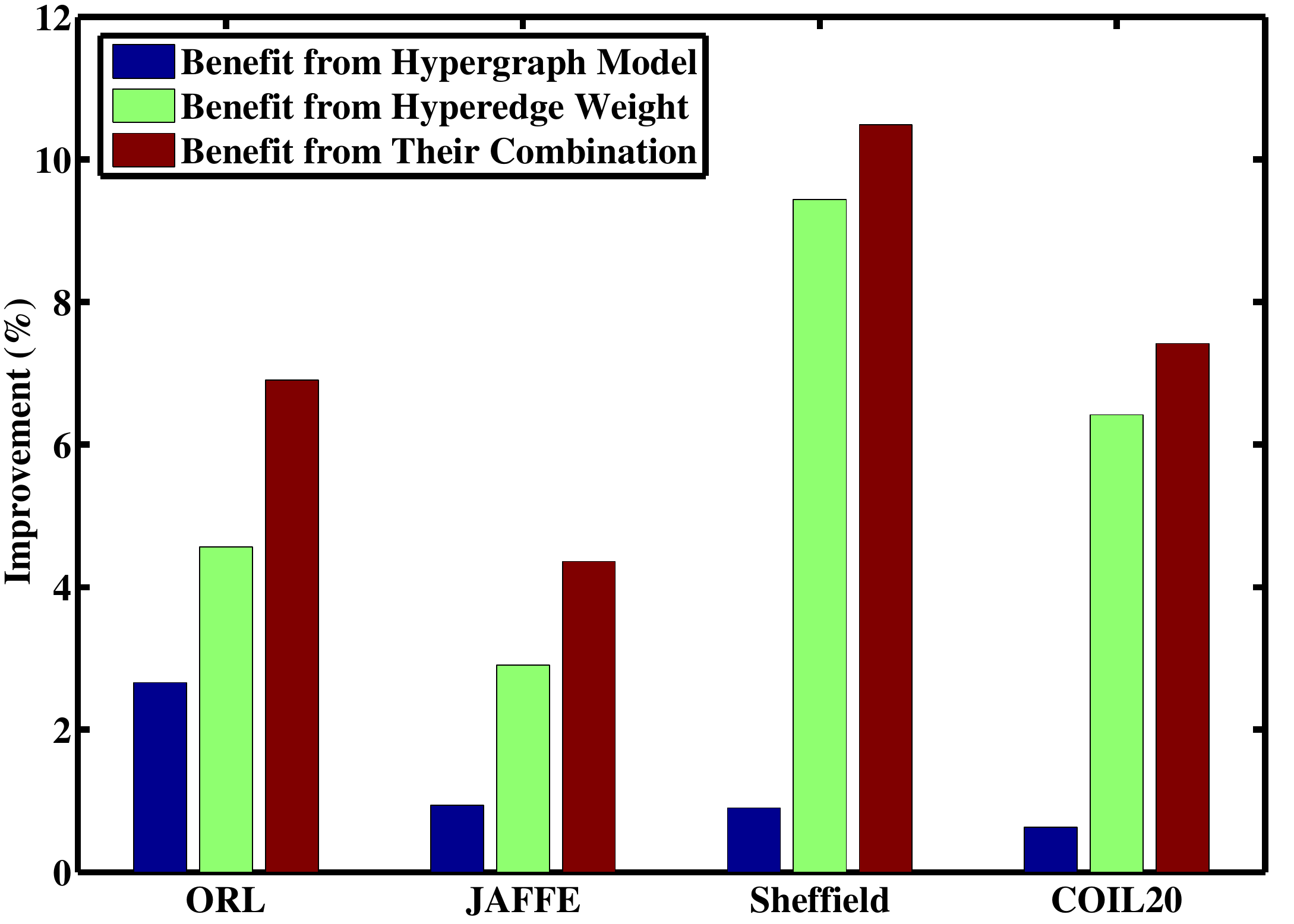}
}
\subfigure[40\% samples for training]{
\centering
\includegraphics[scale=0.25]{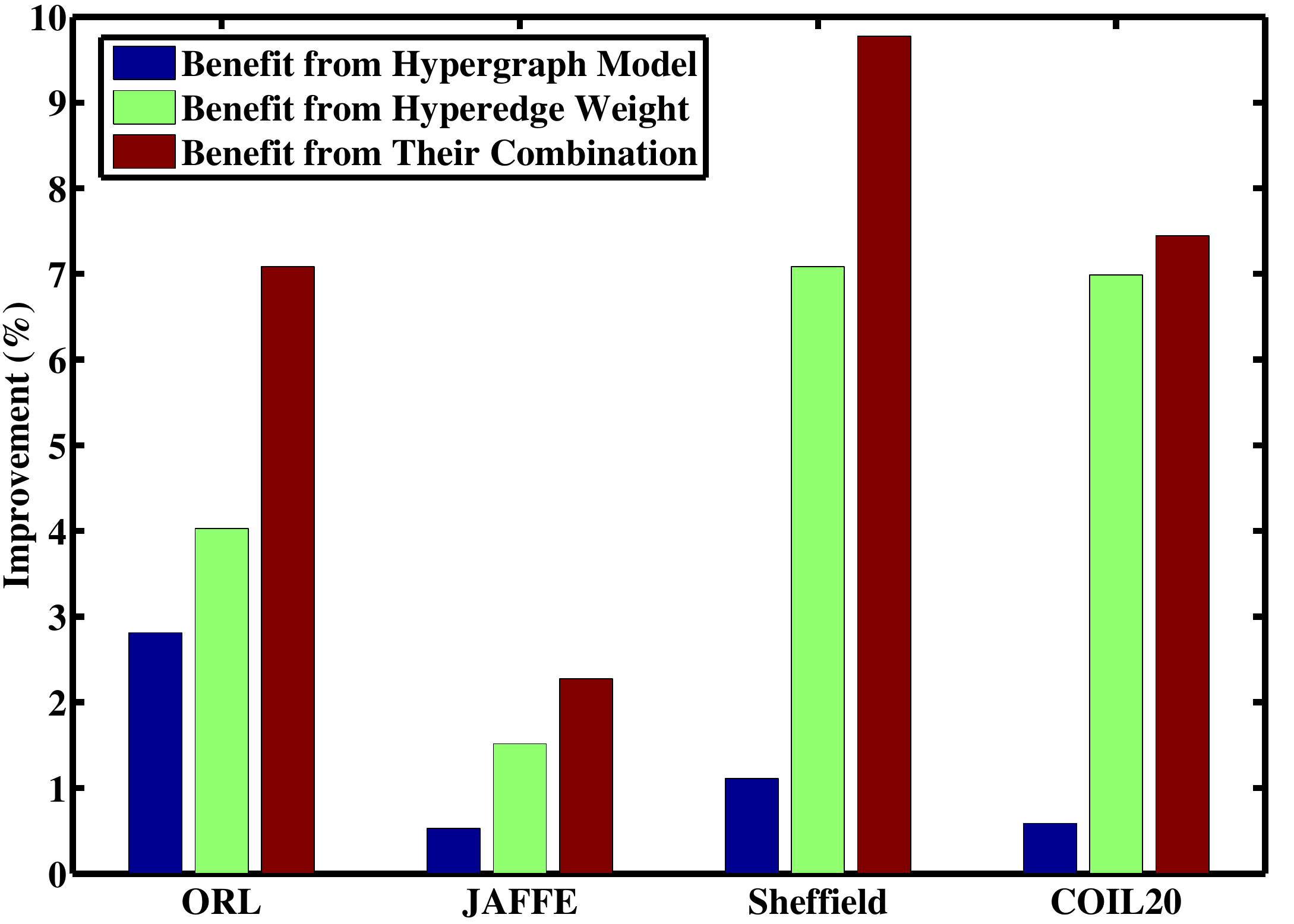}
}
\subfigure[50\% samples for training]{
\centering
\includegraphics[scale=0.25]{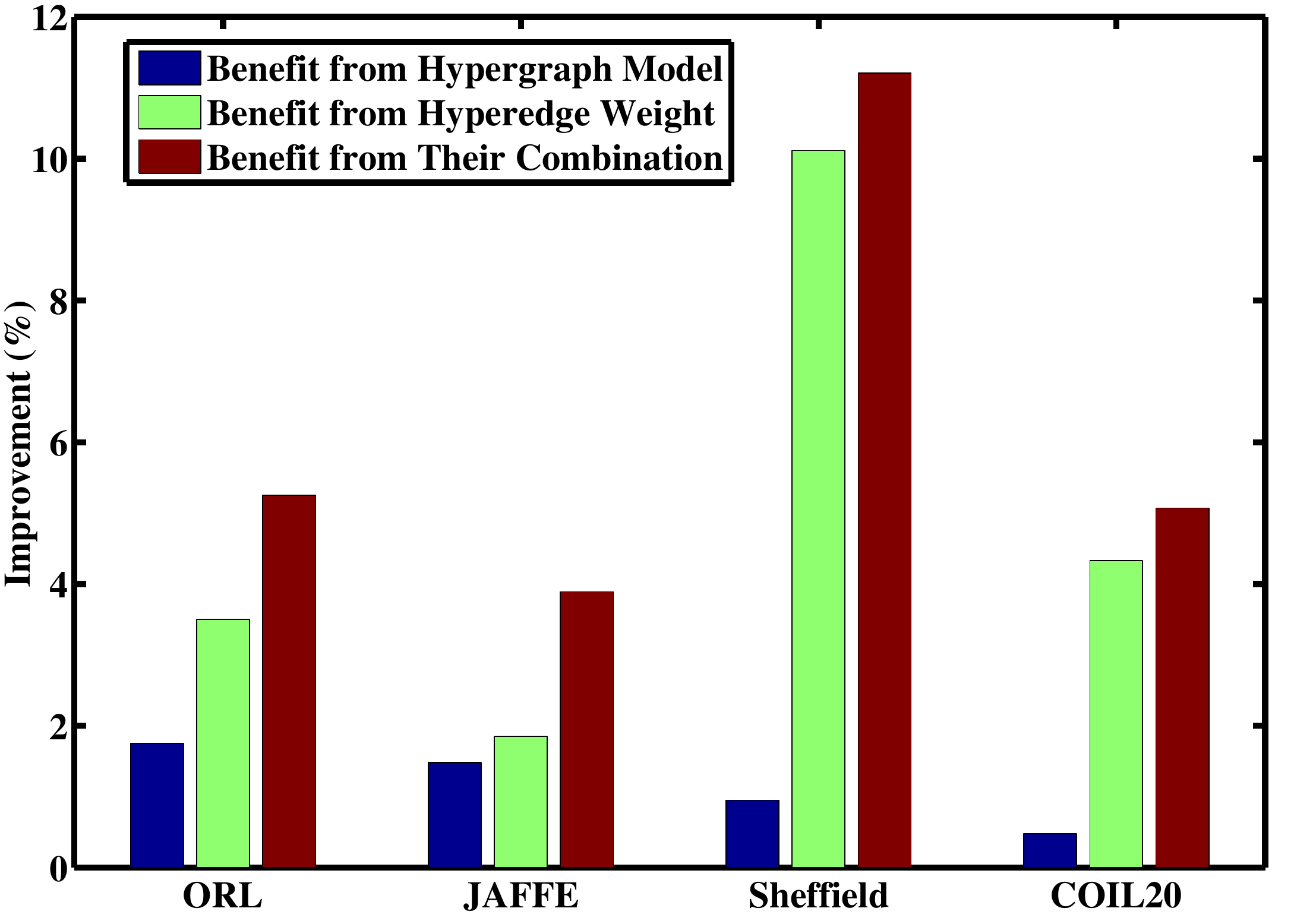}
}
\caption{The classification accuracy improvements of the best choices of hyperedge weight, hypergraph model and their combination, over the worst choices under different training sample perecents.  }
\label{Sen_C}
\end{figure*}

Moreover, several experiments are conducted for studying the influences of the choices of the hypergraph framework versus the choice of hyperedge weighting scheme to the classification performances. To measure
the impact of the choice of the hypergraph framework, we measure the classification accuracy improvement of the best framework choice over the worst choice.
We use the same strategy to measure the impact of the choice of the hyperedge weighting scheme, and the choice of their combination to the classification performance. Figure~\ref{Sen_C} reports the results of these experiments. The results demonstrate that the classification performance is benefited much more from a good choice of hyperedge weight than a good choice of hypergraph framework in all experiments. In the most of cases, the positive impact from a good hyperedge weight is five times even ten times of the positive impact from a good hypergraph framework. This phenomenon reveals the importance of hyperedge weight in hypergraph-based classification.


\subsection{Evaluation in Clustering}
\label{hyperclus}
In this section, we report several experiments that are conducted for studying the influence of hyperedge weight to a clustering task. At first, we apply different hypergraph algorithms to learn the embedding of data. After that, k-means is adopted to predict the label of samples based on the embedding results, and the number of clusters is fixed to the number of classes. The clustering result is evaluated by comparing the predicted label of each sample with the label provided by the data set. Two metrics, the accuracy (AC) and the normalized mutual information metric (NMI) are used to measure the clustering performance. Please see \cite{lpi} for the detailed definitions of these two metrics.
\begin{table*}[h]
\footnotesize
    \caption{Clustering performances of Zhou's Normalized Laplacian using different hyperedge weights on ORL, COIL20, JAFFE, Sheffield databases. (The higher is better) }
    \label{CL_Zhou}
  \begin{center}
   \begin{tabular}{c| p{1.3cm}<{\centering} p{1.3cm}<{\centering} p{1.3cm}<{\centering} p{1.3cm}<{\centering} p{1.3cm}<{\centering} p{1.3cm}<{\centering}}
    \hline
    \multirow{2}*{Database}& \multicolumn{6}{c}{Accuracy (Mutual Information Metric, \%) }\\ \cline{2-7}
     &BINARY& SUM & CENTROID &TRACE& VOLUME &LLRE\\
     \hline
ORL&66.75(82.62)&66.75(81.57)&\textbf{69.50(81.62)}&68.50(\textbf{81.62})&\textbf{69.50(81.62})&67.50(79.96)\\
JAFEE&59.44(65.86)& \textbf{65.00}(65.20)&\textbf{65.00(73.36)}&64.44(72.10)&58.33(67.72)&58.33(67.72)\\
Sheffield&64.35(\textbf{77.77})&\textbf{65.04}(75.82)&64.87(75.82)&64.87(75.82)&64.87(75.82)&64.87(75.82)\\
COIL20& 60.42(74.30)&76.60(85.26)&76.60(85.26)&76.60(85.26)&76.46(85.58)&\textbf{79.86(88.02)}\\
\hline
Average&62.74(75.16)&68.35(76.96)& \textbf{68.99(79.01)}& 68.60(78.7)& 67.29(77.69)& 67.64(77.88)\\
    \hline
    \end{tabular}

  \end{center}
\end{table*}

\begin{table*}[h]
\footnotesize
    \caption{Clustering performances of Clique Expansion using different hyperedge weights on ORL, COIL20, JAFFE, Sheffield databases. (The higher is better)}
    \label{CL_Clique}
  \begin{center}
   \begin{tabular}{c|p{1.3cm}<{\centering} p{1.3cm}<{\centering} p{1.3cm}<{\centering} p{1.3cm}<{\centering} p{1.3cm}<{\centering} p{1.3cm}<{\centering}}
    \hline
    \multirow{2}*{Database}& \multicolumn{6}{c}{Accuracy (Mutual Information Metric, \%) }\\ \cline{2-7}
     & BINARY & SUM  & CENTROID &TRACE& VOLUME &LLRE\\
     \hline
ORL&65.25(82.20)& 67.50(80.98)&65.75(82.9)&\textbf{70.50}(82.75)&65.00(82.44)&65.00(80.76)\\
JAFEE&71.11(70.16)&70.00(71.62)&69.44(68.71)&71.67(72.87)&\textbf{74.44(76.27)}&71.11(69.26)\\
Sheffield&61.91(78.13)&\textbf{66.09}(80.47)&\textbf{66.09}(80.47)&\textbf{66.09}(80.47)&\textbf{66.09}(80.47)&\textbf{66.09(80.69)}\\
COIL20&69.72(79.82)&\textbf{82.29(89.12)}&79.31(89.00)&\textbf{82.29(89.12)}&\textbf{82.29(89.12)}&\textbf{82.29(89.12)}\\
\hline
Average&67.00(77.56)&71.47(80.55)& 70.15(80.27)& \textbf{72.64}(81.30)&71.96(\textbf{82.08})& 71.12(79.96)\\
    \hline
    \end{tabular}

  \end{center}
\end{table*}

\begin{table*}[!tbp]
\footnotesize
    \caption{Clustering performances of Star Expansion using different hyperedge weights on ORL, COIL20, JAFFE, Sheffield databases. (The higher is better)}
    \label{CL_Star}
  \begin{center}
   \begin{tabular}{c| p{1.3cm}<{\centering}  p{1.3cm}<{\centering} p{1.3cm}<{\centering} p{1.3cm}<{\centering} p{1.3cm}<{\centering} p{1.3cm}<{\centering}}
    \hline
    \multirow{2}*{Database}& \multicolumn{6}{c}{Accuracy (Mutual Information Metric, \%) }\\ \cline{2-7}
     & BINARY & SUM & CENTROID &TRACE& VOLUME &LLRE\\
     \hline
ORL&63.75(79.01)&65.75(79.48)&65.75(79.48)&67.00(80.48)&65.75(79.48)&\textbf{68.25(80.52)}\\
JAFEE&61.11(67.51)&63.33(70.23)&61.11(67.51)&\textbf{65.56(70.54)}&61.67(68.16)&63.89(69.23)\\
Sheffield&58.43(72.7)&\textbf{62.96}(75.10)&62.43(\textbf{77.62})&62.26(73.73)&58.26(74.28)&58.26(73.89)\\
COIL20&49.65(64.67)&62.36(73.15)&62.43(\textbf{73.86)}&56.46(71.29)&59.03(72.53)&\textbf{62.99}(73.62)\\
\hline
Average&58.23(70.97)&\textbf{63.60(74.49)}&62.93(74.62)&62.82(74.01)&61.18(73.61)&63.35(74.31)\\
    \hline
    \end{tabular}

  \end{center}
\end{table*}

\begin{figure*}[h]
\subfigure[Accuracy]{
\centering
\includegraphics[scale=0.28]{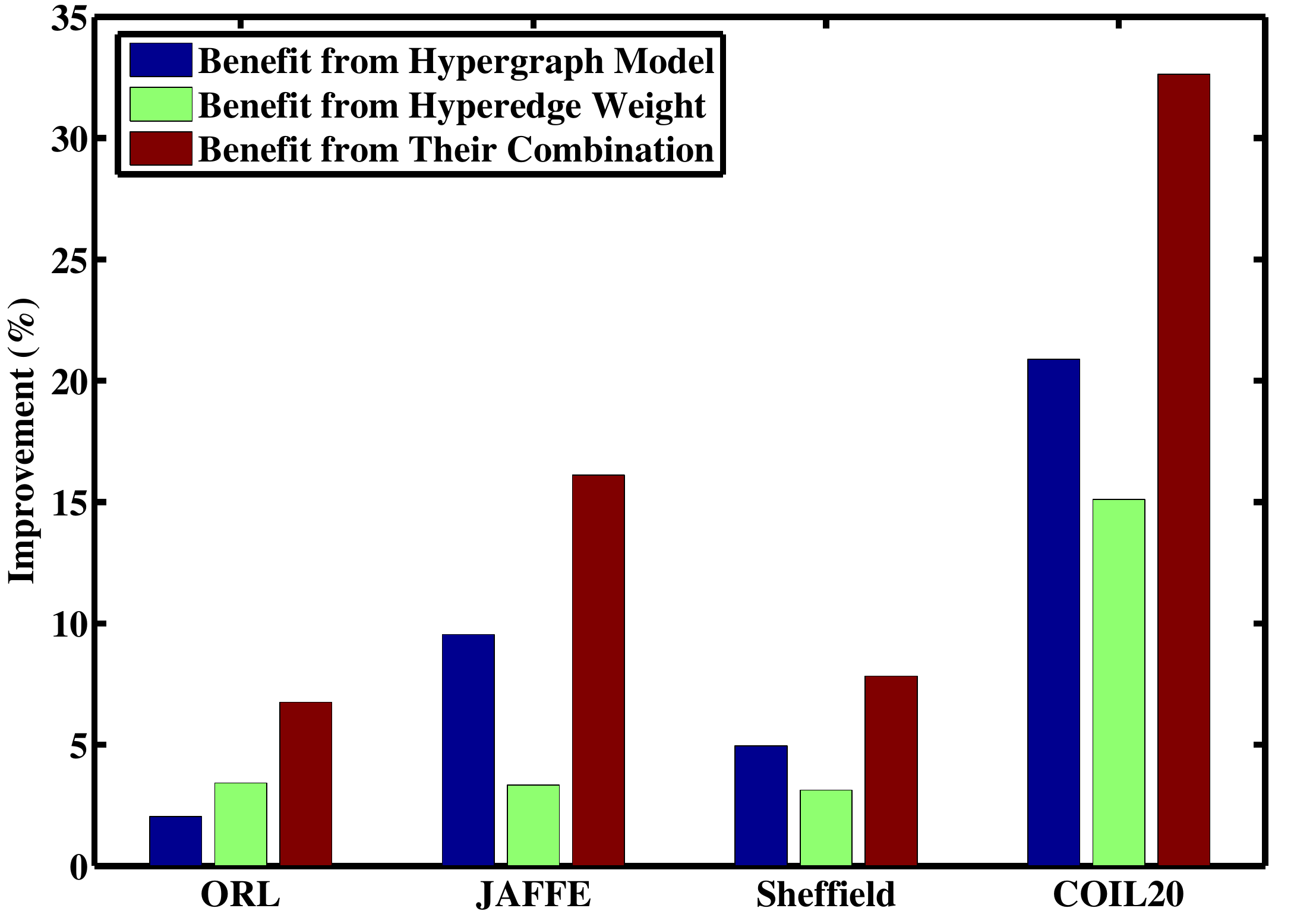}
}
\subfigure[NMI]{
\centering
\includegraphics[scale=0.28]{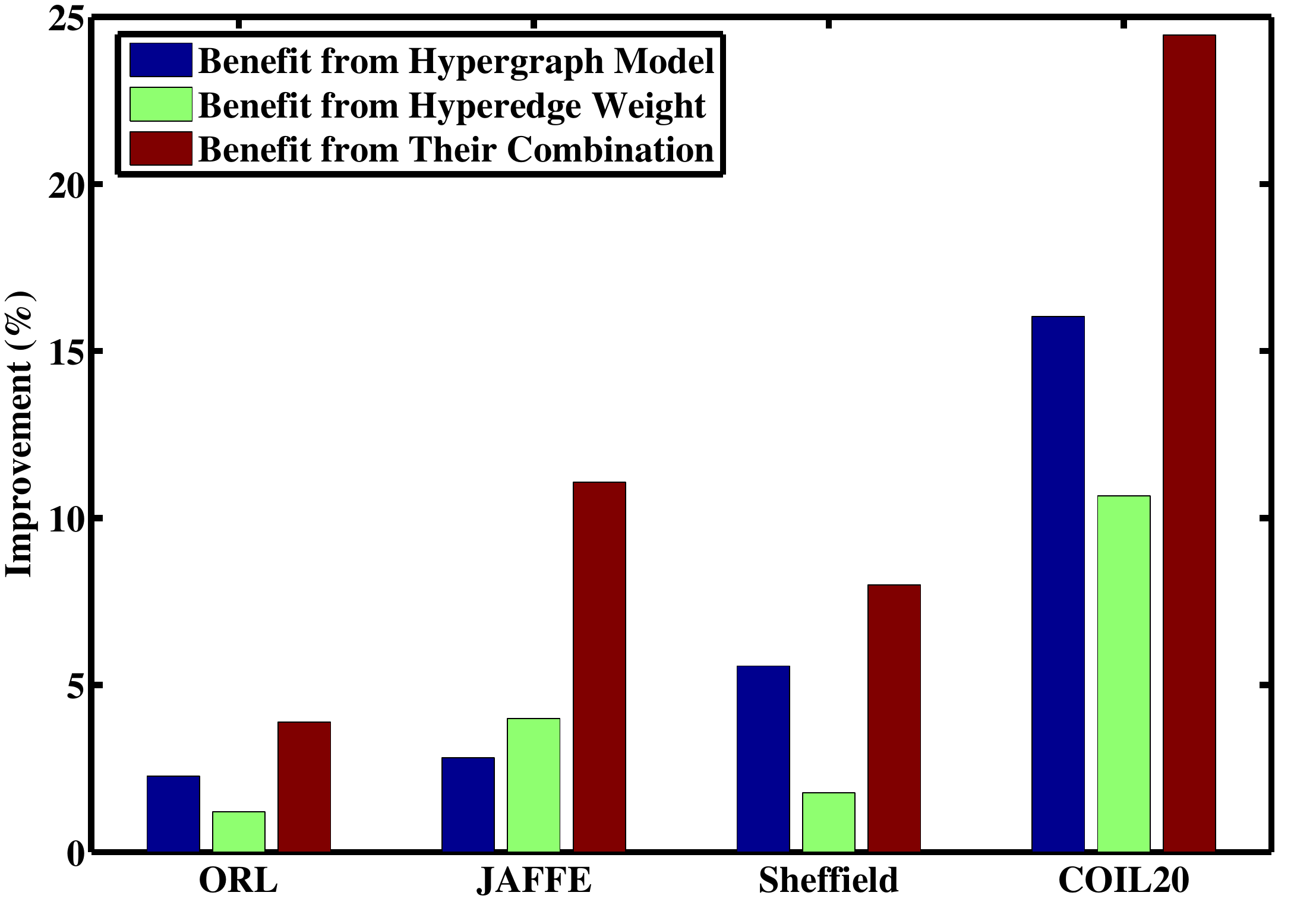}
}
\caption{The average classification accuracy improvements of the best choices of hyperedge weight, hypergraph model and their combination, over their worst choices on four databases.  }
\label{cluster_evaluate}
\end{figure*}

Tables \ref{CL_Zhou}, \ref{CL_Clique} and \ref{CL_Star} respectively report the clustering results of Zhou's Normalized Laplacian, Clique Expansion and Star Expansion frameworks using different hyperedge weights on four databases. From the experimental results, it seems that different weighting schemes performs well on different databases. From the comprehensive evaluation based on the mean accuracy and mean mutual information metric of four databases, CENTRIOD and TRACE slightly performs better than VOLUME, SUM and LLRE. However, another interesting phenomenon is that all five weighting schemes significantly outperform the unweighted case, BINARY. According to these observations, we conclude that the hyperedge weight still plays an important role in hypergraph-based clustering. However unlike classification case, the hypergraph-based clustering is not so sensitive to the choice of hyperedge weight. We follow the same evaluation strategy in the Section~\ref{hyperclass} to study the influences of the choices of the hypergraph framework and the hyperedge weighting scheme to the clustering performances. Figure~\ref{cluster_evaluate} shows the experimental results under two different evaluation metrics. In the most of case, hypergraph-based clustering can benefit more from a good choice of hypergraph framework than a good choice of hyperedge weight. But, these two benefits are in the same level. So, we still cannot ignore the positive influence of a good hyperedge weight for clustering.

\subsection{The Setting of Scaling Parameter $\mu$}
\label{mu}
 In this section, we conduct several experiments to study the influence of the scaling parameter $\mu$ to the performances of hypergraph learning and also experimentally find optimal $\mu$ on ORL, COIL20, Sheffield and JAFFE databases. We conduct the experiments for addressing the clustering the tasks. The learned optimal $\mu$ is directly applied to hypergraph-based classification, which has already been introduced in Section~\ref{hyperclass}. The experimental settings are the same as the settings of hypergraph-based clustering introduced in Section~\ref{hyperclus}. We normalize all the hyperedge weights via dividing by the mean of the hyperedge weights before exponentiation, $w_i=exp(-(w_i/\bar{w})/\mu)$ where $\bar{w}$ is the mean of hyperedge weights. After this operation, the different hyperedge weight schemes are in the same scale. Since there are two evaluation metrics for clustering, we use the mean of these two metrics as a new metric to evaluate the clustering performance and then find the optimal $\mu$ for each weighting scheme.

\begin{figure*}[!h]

\subfigure[ZNH on ORL ]{
\centering
\includegraphics[scale=0.19]{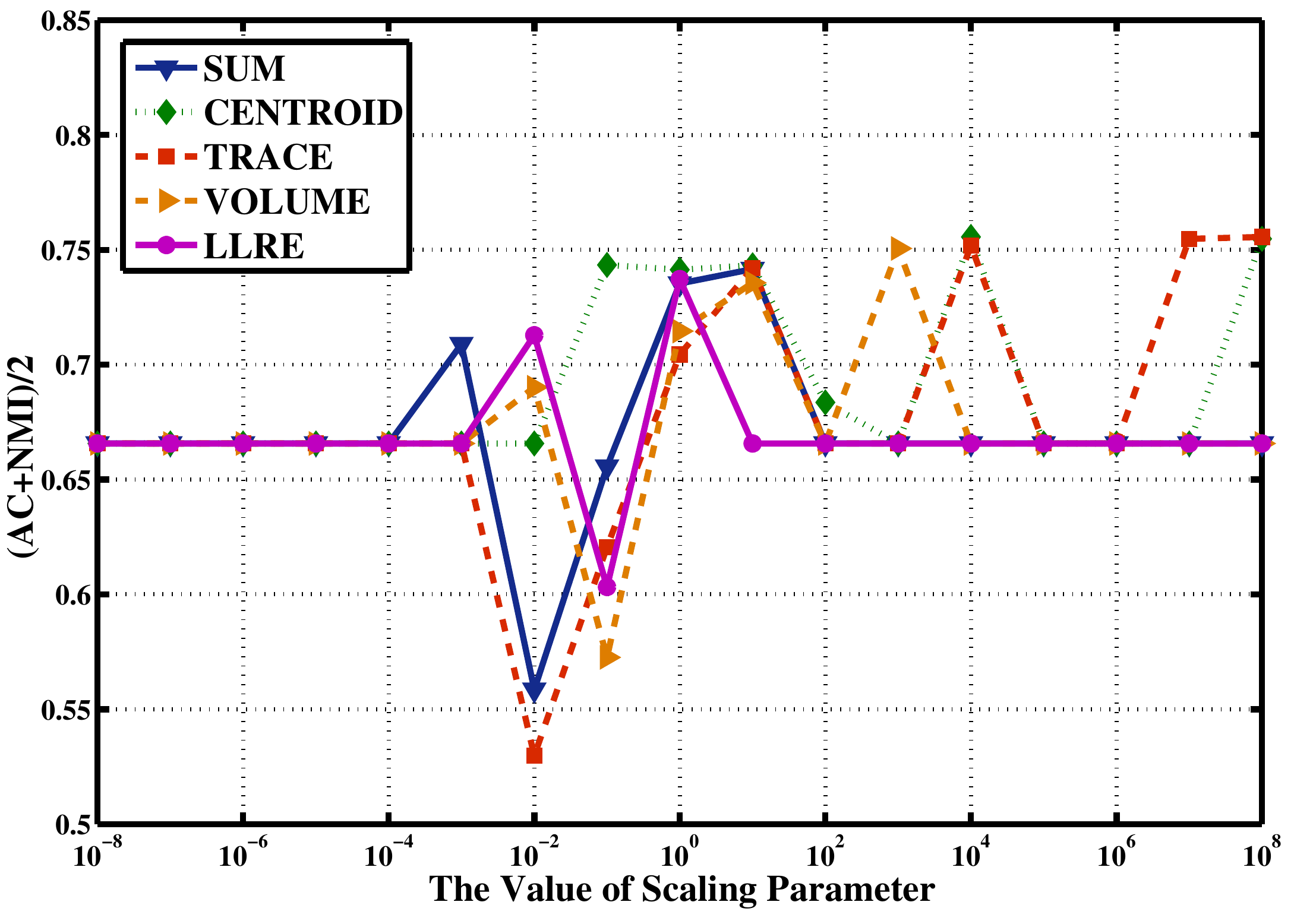}
}
\subfigure[CliqueExp on ORL ]{
\centering
\includegraphics[scale=0.19]{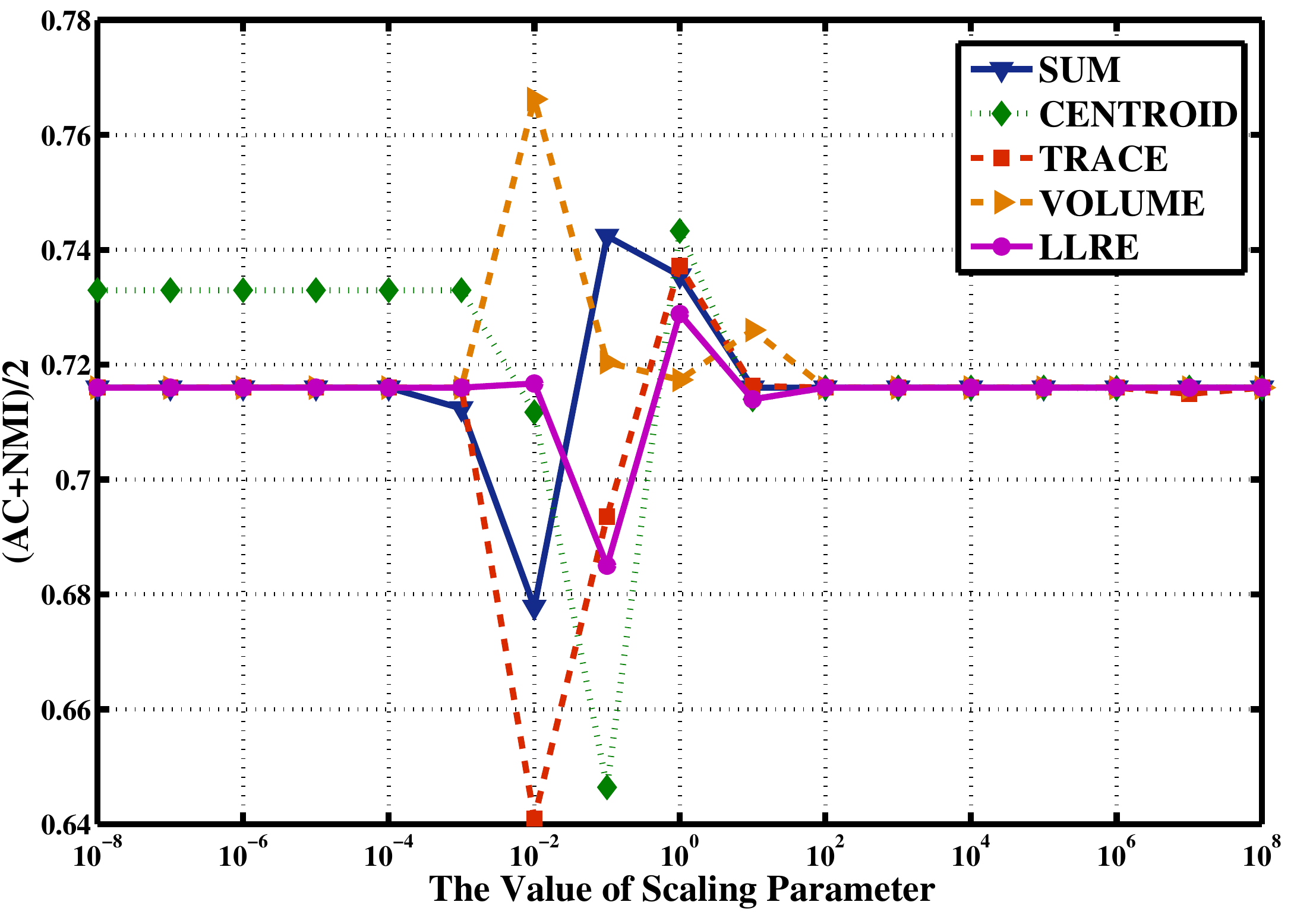}
}
\subfigure[StarExp on ORL]{
\centering
\includegraphics[scale=0.19]{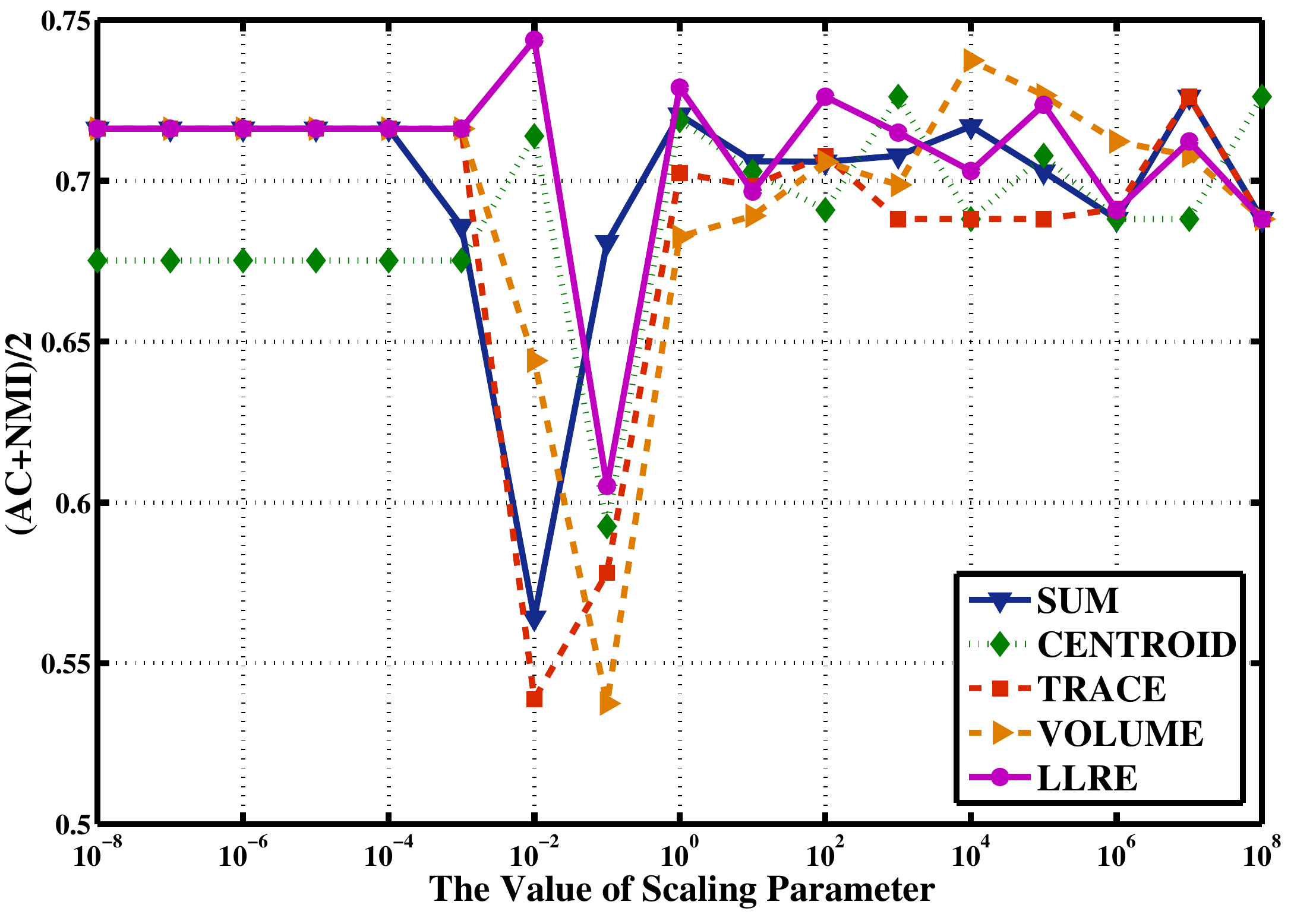}
}
\subfigure[ZNH on JAFFE ]{
\centering
\includegraphics[scale=0.19]{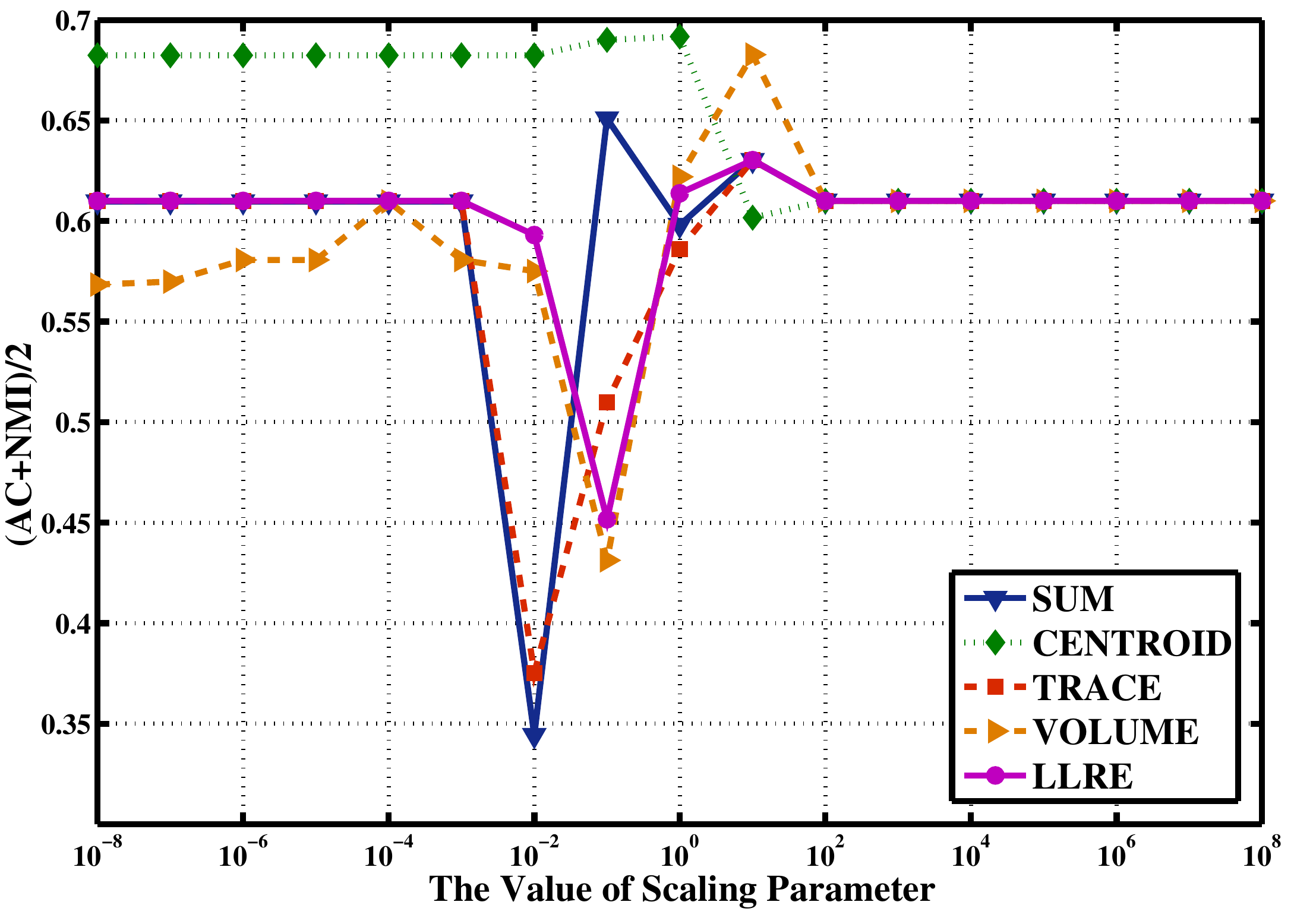}
}
\subfigure[CliqueExp on JAFFE ]{
\centering
\includegraphics[scale=0.19]{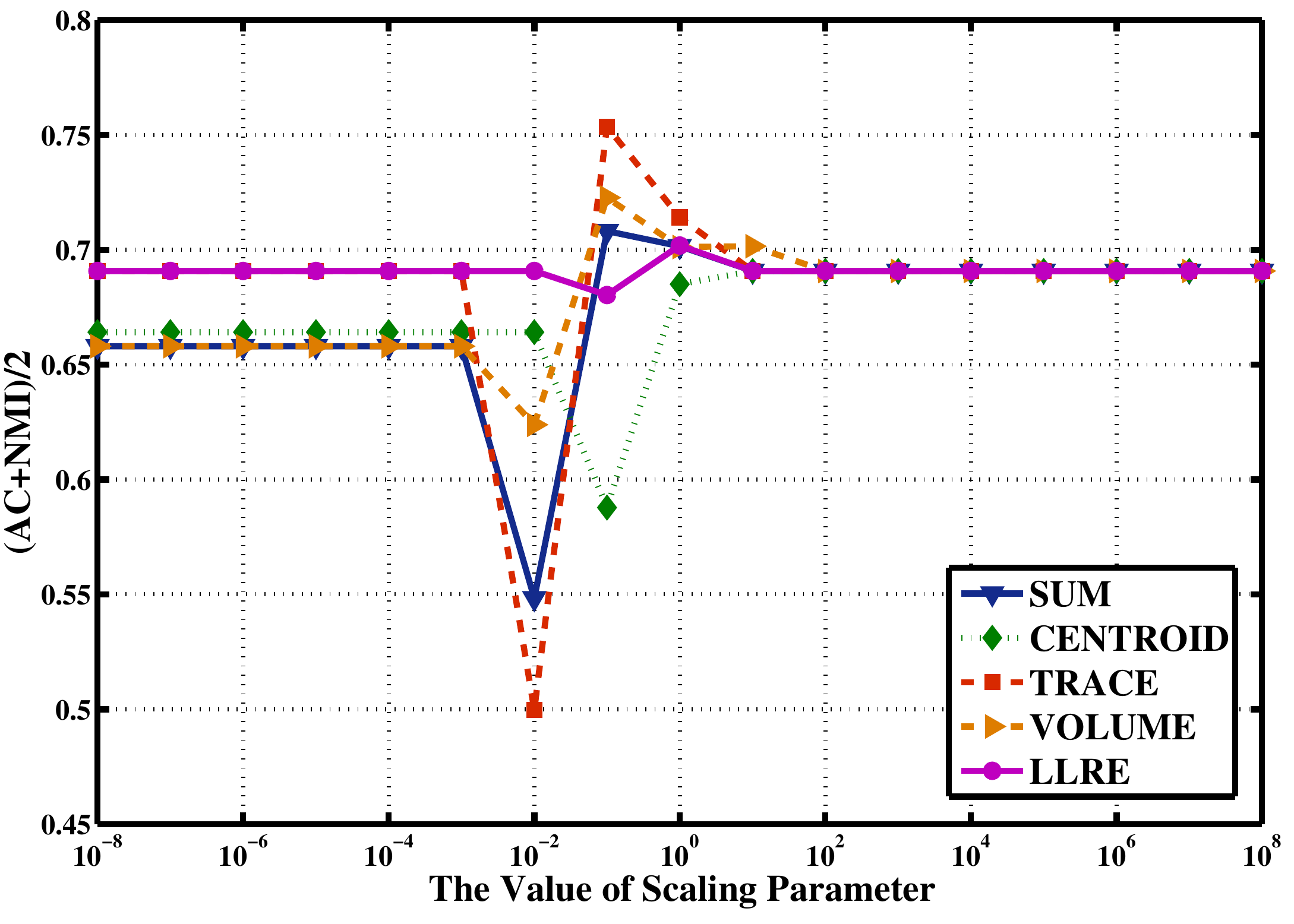}
}
\subfigure[StarExp on JAFFE]{
\centering
\includegraphics[scale=0.19]{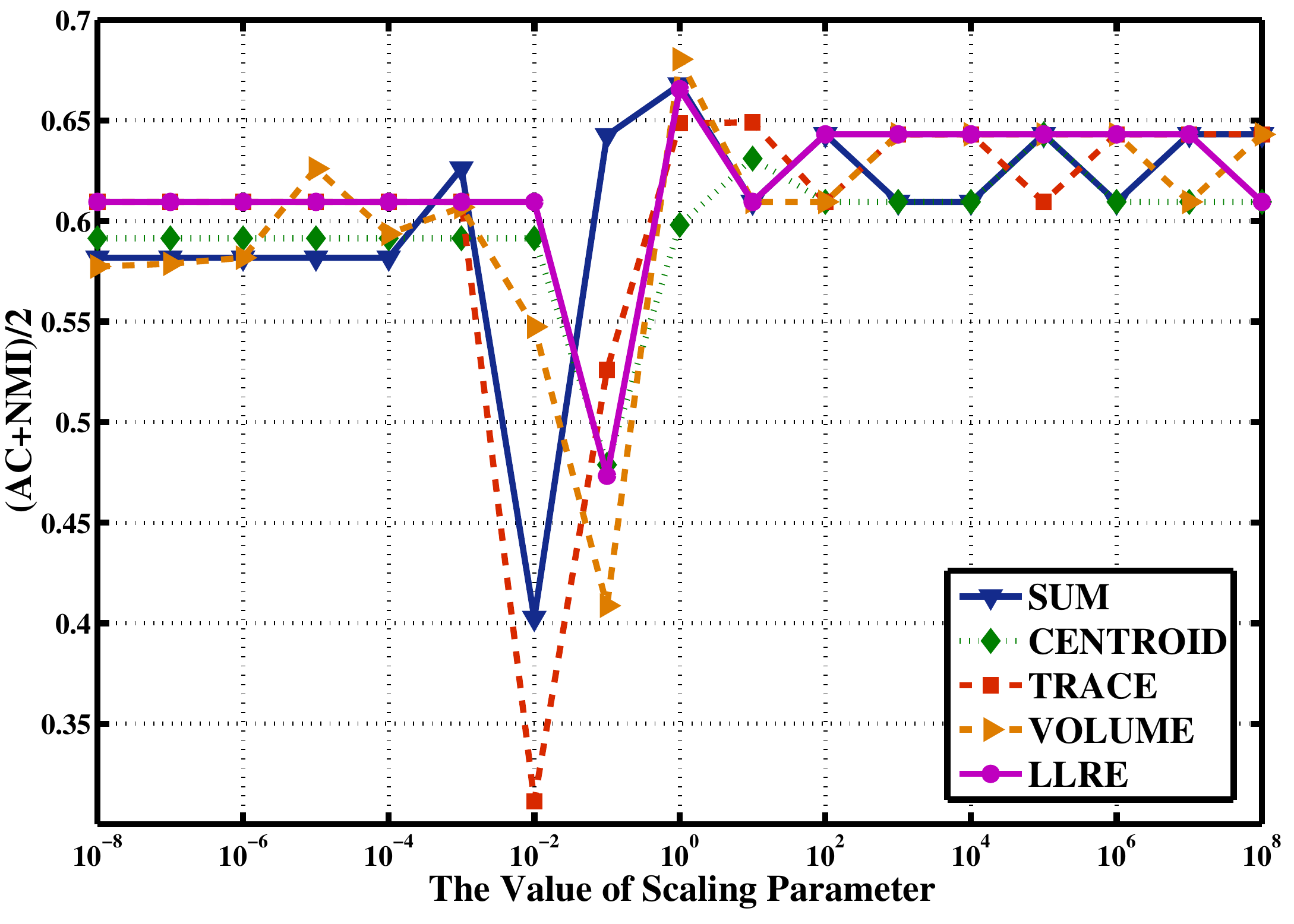}
}
\subfigure[ZNH on Sheffield ]{
\centering
\includegraphics[scale=0.19]{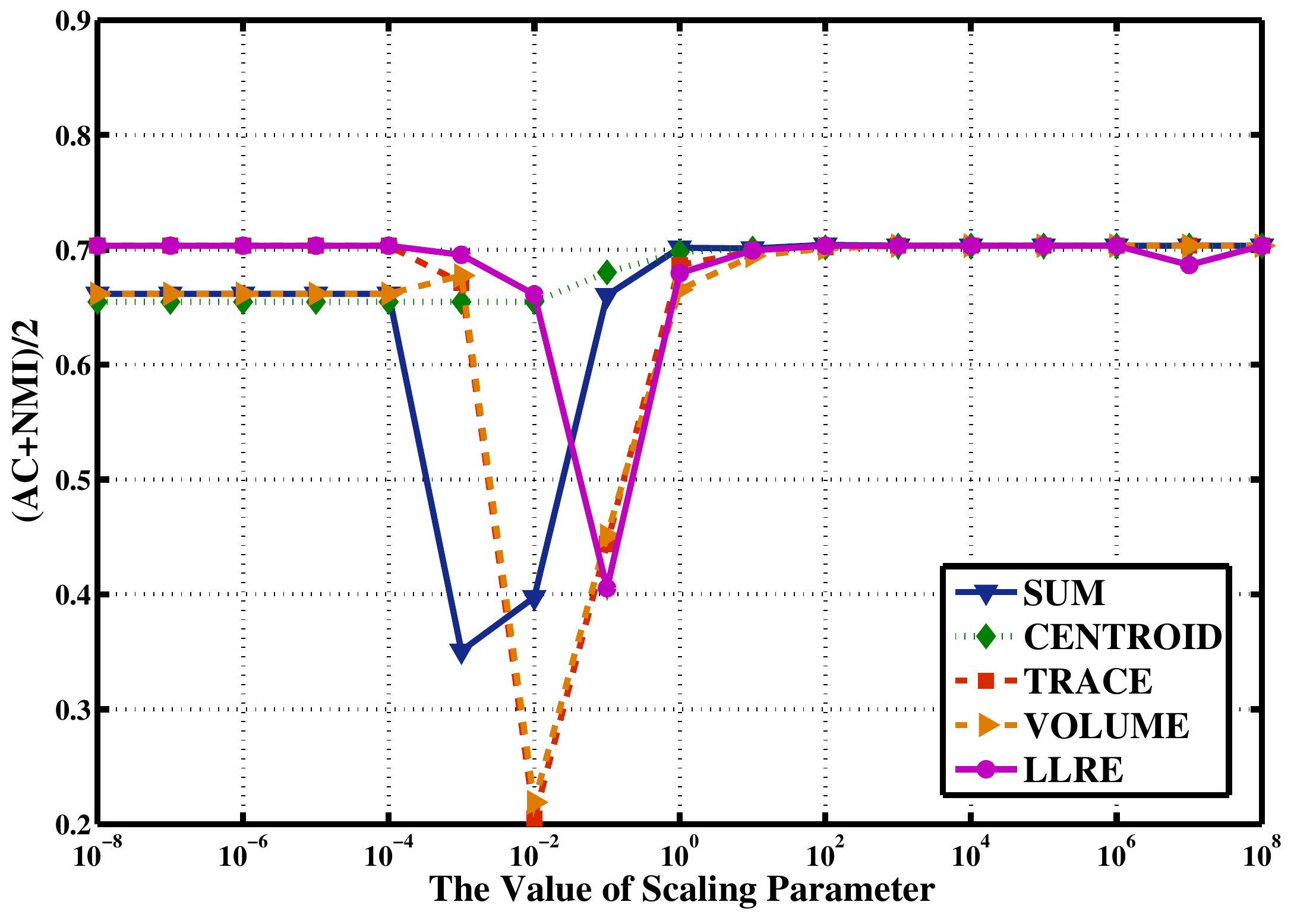}
}
\subfigure[CliqueExp on Sheffield ]{
\centering
\includegraphics[scale=0.19]{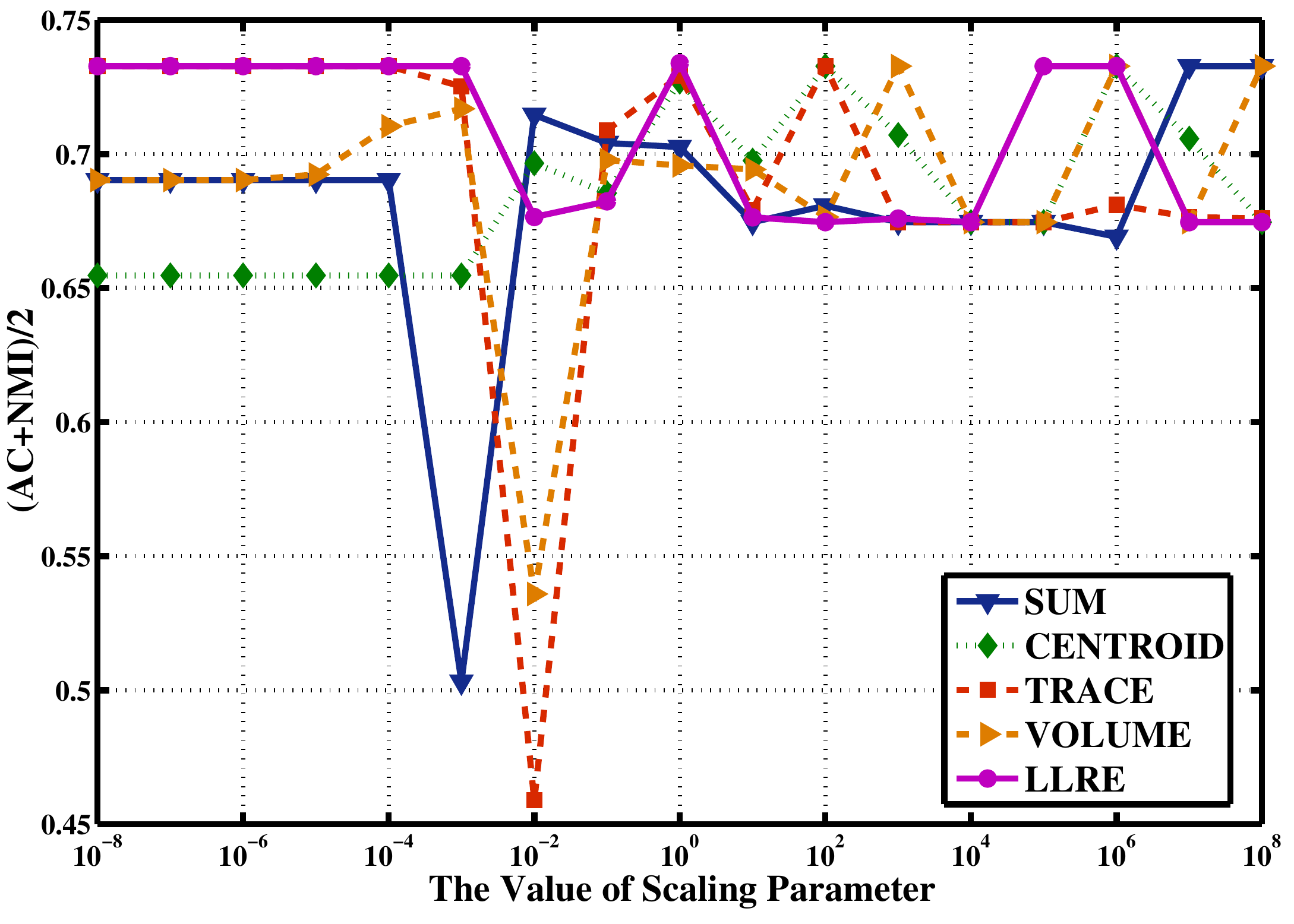}
}
\subfigure[StarExp on Sheffield]{
\centering
\includegraphics[scale=0.19]{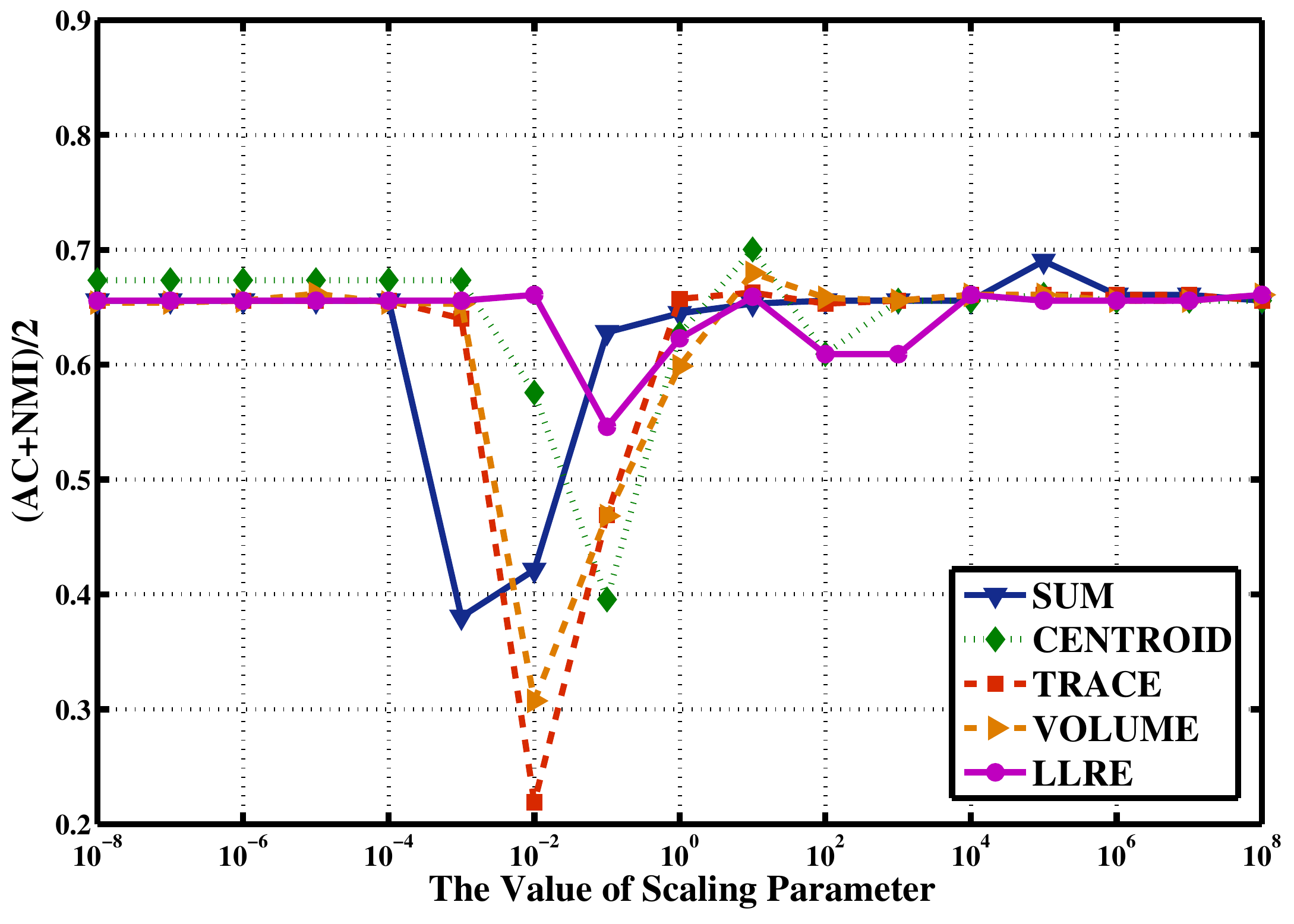}
}
\subfigure[ZNH on COIL20 ]{
\centering
\includegraphics[scale=0.195]{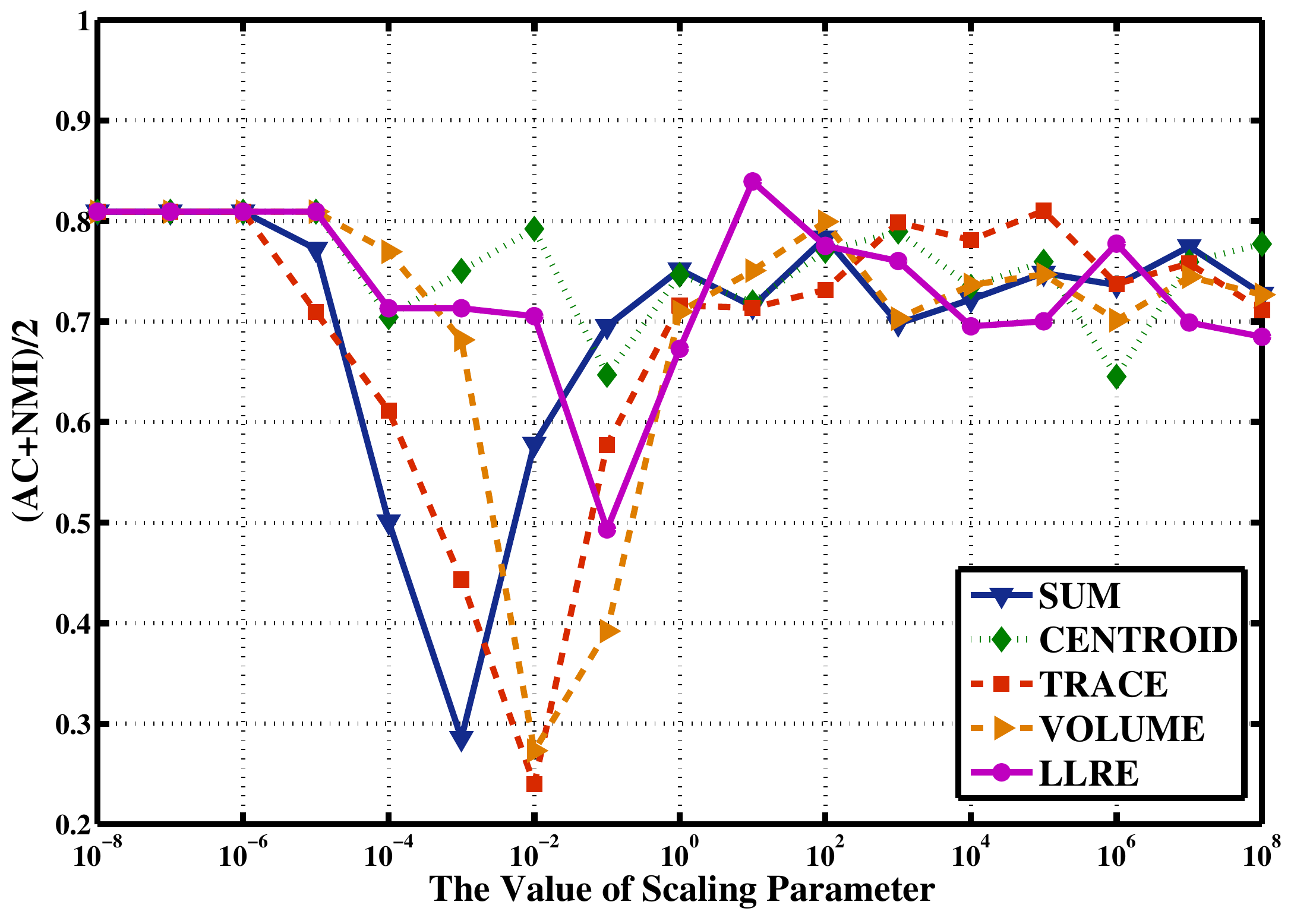}
}
\subfigure[CliqueExp on COIL20 ]{
\centering
\includegraphics[scale=0.195]{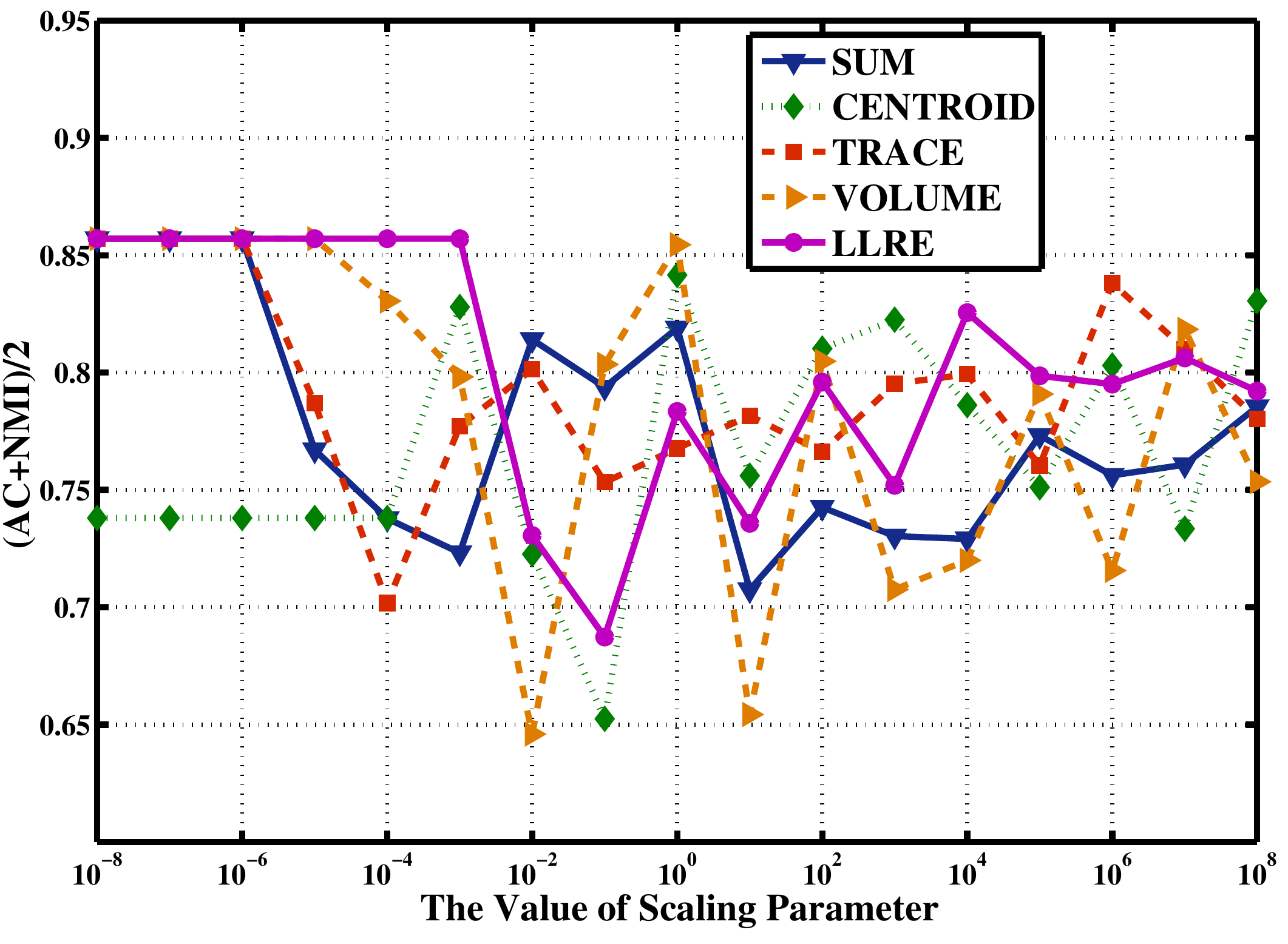}
}
\subfigure[StarExp on COIL20]{
\centering
\includegraphics[scale=0.195]{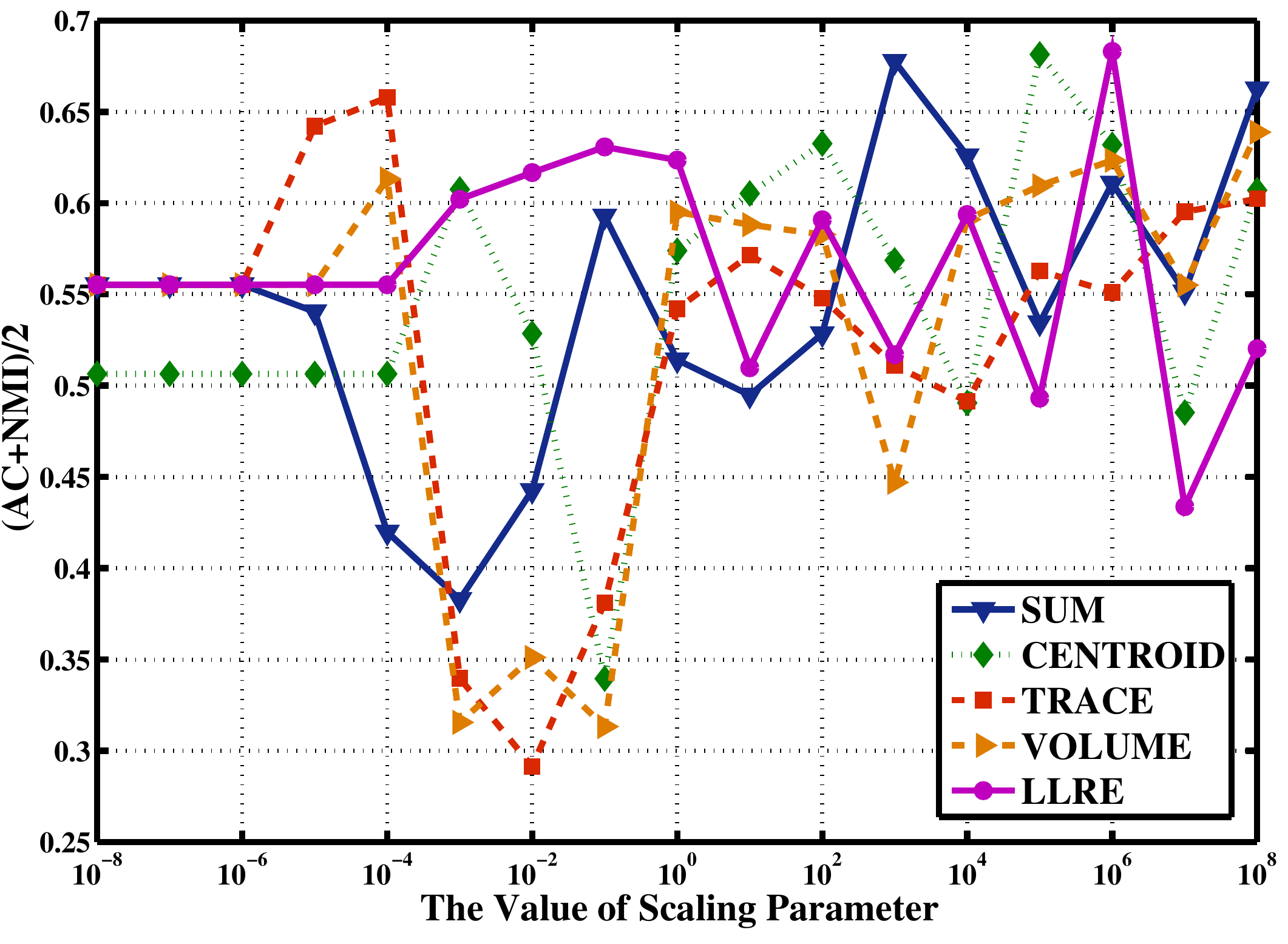}
}
\caption{The effect of $\mu$ for clustering on ORL, JAFFE, Sheffield and COIL20 databases using hyperedge weights of SUM, CENTROID, TRACE, VOLUEM and LLRE. (ZNH = Zhou's Normalized Hypergraph, CliqueExp = Clique Expansion, StarExp = Star Expansion)  }
\label{optu}
\end{figure*}

Figure~\ref{optu} shows the experimental results. From the observations of Figure~\ref{optu},  it seems that the optimal $\mu$ is more related to the database instead of the hypergraph model. In the most of case, CENTROID, TRACE and LLRE can obtain a reasonable performance when $\mu$ is equal to 1 or 10. SUM and VOLUME are more sensitive to the $\mu$, it is hard to find an optimal $\mu$ that works for all database. Anyway, a good $\mu$ can be learned from training set. This is a commonly adopted strategy in many practical applications.

\subsection{Comparison with The State-of-the-Arts}
To further show the value of our work, we select the best performed weighting schemes based on the results of the prior experiments and apply them to both image classification and data representation for clustering. TRACE and VOLUME-based Zhou's Normalized Hypergraph are used for classification while CENTORID-based Zhou's Normalized Hypergraph and TRACE-based Clique Expansion are adopted for clustering. In these experiments, two challenging databases named Scene15 and Caltech256 databases are added. Adaptive-Hypergraph (Ada-Hyper) \cite{adaptive}, Relaxed Collaborative Representation classifier (RCR)~\cite{rcr}, Sparse Representation Classifier (SRC)~\cite{sparse}, Random Forest classifier (RF)\cite{rf} and graph-based classifier are the compared methods for classifications.

Graph Regularized Nonnegative Matrix Factorization (GNMF) \cite{gnmf}, Landmark-Based Spectral Clustering (LSC)~\cite{lbsc}, Sparse Representation-based Embedding (SRE)~\cite{sgraph}, Graph-based Normalized cut~\cite{ncut} (or Laplacian EigenMapping~\cite{eigenmap}) and K-means are the compared algorithms for clustering. All the classification experiments are conducted in two-fold cross-validation case. With regard to the clustering experiments, the numbers of clusters are all set as the number of classes for each database. The experimental settings of image classification are following \cite{adaptive} while the experimental settings of the clustering task is following \cite{gnmf}. The choices of $k$ in image classification experiments are all following the same setting in the previous experiments. With regard to the image clustering, we find that a combination of $k$ is much more suitable to the challenging databases, Scene15 and Caltech256 databases. So we let $k=[10,20,30,40,50]$ on Scene15 dataset, and $k=[50,100]$ on Caltech256 dataset. With regard to ORL and COIL20 datasets, we follow the same setting in Section~\ref{hyperclus}.

\begin{table*}[!tbp]
\footnotesize
  \begin{center}
      \caption{Image classification performances of five classifiers. (ZNH = Zhou's Normalized Hypergraph)}
   \label{C_classify}
    \begin{tabular}{c p{1.4cm}<{\centering} p{1.4cm}<{\centering} p{1.4cm}<{\centering} p{2cm}<{\centering}}
    \hline
    \multirow{2}*{Database}& \multicolumn{4}{c}{Mean Classification Errors $\pm$ Standard deviation (\%)}\\ \cline{2-5}
    &ORL & COIL20 & Scene15 & Caltech256-2000\\
    \cline{1-5}
    ZNH+TRACE&5.50$\pm$1.41&\textbf{3.68$\pm$0.69}&25.47$\pm$1.89&42.75$\pm$2.90\\
    ZNH+VOLUME&\textbf{4.50$\pm$1.41}&4.79$\pm$0.29&\textbf{25.40$\pm$1.98}&42.70$\pm$2.97\\
    Graph&24.75$\pm$1.05&43.13$\pm$0.49&32.40$\pm$2.07&51.35$\pm$3.75\\
    Adap-Hyper\cite{adaptive} &11.81$\pm$0.42&10.13$\pm$1.41& 25.54$\pm$1.59&43.30$\pm$2.10\\
    RCR\cite{rcr} &8.00$\pm$2.07&11.19$\pm$1.10&26.73$\pm$1.43&\textbf{41.21$\pm$2.11}\\
    Random Forest~\cite{rf}&11.54$\pm$2.13&6.00$\pm$0.10&27.66$\pm$2.63&43.74$\pm$3.02\\
    SRC~\cite{sparse}&8.25$\pm$1.06&11.81$\pm$0.98&26.80$\pm$2.83&42.80$\pm$1.41\\
    \hline
    \end{tabular}

  \end{center}
\end{table*}

\begin{table*}[!tbp]
  \begin{center}
  \footnotesize
      \caption{Clustering performances of five algorithms. (CliqueExp = Clique Expansion) }
    \label{C_cluster}
    \begin{tabular}{c p{1.5cm}<{\centering} p{1.5cm}<{\centering} p{1.5cm}<{\centering} p{2cm}<{\centering}}
    \hline
    \multirow{2}*{Database}& \multicolumn{4}{c}{Clustering Accuracy (Normalized Mutual Information, \%)}\\ \cline{2-5}
    &ORL & COIL20 & Scene15 & Caltech256-2000\\
    \cline{1-5}
    CliqueExp+TRACE&\textbf{70.50}(82.75)&\textbf{82.29}(89.12)&\textbf{67.47}(62.03)&36.30(34.01)\\
    ZNH+CENTROID&69.50(81.62)&76.60(85.26)&63.33(59.53)&\textbf{44.15(43.29)}\\
    GNMF~\cite{gnmf}&65.75(82.19)&82.22\textbf{(89.99)}&62.87(61.32)&39.80(38.45)\\
    LSC~\cite{lbsc}& 66.00(82.39)&76.04(86.13)&66.33(\textbf{64.01})&43.95(41.32)\\
    Graph~\cite{ncut,eigenmap}& 67.75(82.01)&69.60(77.00)&56.87(60.21) &38.00(38.69) \\
    SRE~\cite{sgraph}&70.00(\textbf{83.10})&69.24(76.16)&58.27(59.09)&35.70(38.91)\\
    Kmeans&57.75(78.38)&63.70(73.40)&63.20(62.45) &41.15(40.55) \\
    \hline
    \end{tabular}

  \end{center}
\end{table*}

Table~\ref{C_classify} reports the classification performances of five different classifiers and two hypergraph-based classifiers. It is clear that the VOLUME and TRACE-based hypergraph classifiers outperforms all other classifiers on ORL, COIL20 and Scene15 databases and respectively get the second and third places on Clatech256 database. Compared to the regular pairwise graph-based classifier, the gains of the VOLUME-based hypergraph classifier are 20.45\%, 38.34\%, 7.00\% and 8.65\% on ORL, COIL20, Scene15 and Caltech256 databases respectively while these numbers of the TRACE-based hypergraph classifier are 19.45\%, 39.45\%, 6.93\% and 8.60\%. Even comparing with the state-of-the-art classifiers, such as random forest and sparse representation classifier, our proposed hyperedge weight-based hypergraph classifiers still show their superiorities. Table~\ref{C_cluster} shows the clustering results of the different algorithms. From the results, we can see that TRACE and CENTROID-based hypergraph models can get the best clustering accuracies on all four databases and they also can get the promising NMIs in comparison with the state-of-the-art algorithms. Similar to the results of classification, the advantage of hypegraph models over graph model is very obvious. For exmaple, the clustering accuracy gains of the TRACE-based hypergraph over graph are 2.75\%, 12.69\% and 10.60\% on ORL, COIL20 and Scene15 databases respectively. The experiments verify that a good hypergraph framework with a carefully chosen hyperedge weight is very competitive for classification and clustering. Moreover, it is still possible to further improve the performances of these hypergraphs by using our proposed weighing schemes, since many other settings of the proposed weighting schemes are not explored yet. For example, VOLUME computes the hyperedge weight from edge weights. But, in these experiments, we only adopt the Heat-Kernel weighting scheme to compute edge weight. There are still many other edge weighting schemes can be applied. Similarly, LLRE only uses the common linear regression to compute the local linear reconstruction errors. The more advanced linear regression methods, such as Sparse representation and collaborative representation, have not been tried yet.

\subsection{Experimental Analysis and Discussion}
Several conclusions can be made from the experimental results listed in Figures \ref{overall_C} - \ref{optu} and Tables \ref{C_Zhou} - \ref{C_cluster}. We believe these conclusions are very instructive to the researchers who work at hypergraph learning. Here we give these conclusions:
\begin{enumerate}
 \setlength{\itemsep}{0pt}
 \setlength{\parskip}{0pt}
 \setlength{\parsep}{0pt}
 \item Similar to the choice of the hypergraph algorithm itself, the choice of the hyperedge weight also plays a very important, even a more important role. A prominent improvement can be obtained by carefully choosing the suitable hyperedge weight. This can be noticed widely in our experiments, particular in the experiments of classification. 
 \item Hypergraph model is more sensitive to the hyperedge weight when it is used for classification than for clustering.
 \item The proposed hyperedge weights, VOLUME and TRACE, can be deemed as two representative hyperedge weight schemes for classification, since they distinctly outperform other weight schemes in all experiments. However, for the clustering task, it is hard to conclude a representative weighting scheme, since all five weighting schemes have a similar comprehensive performance. Comparatively speaking, TRACE and CENTROID performs slightly better. Moreover, in the clustering case, all five weighing schemes significantly outperforms the unweighted case.
 \item According to our experimental study, we respectively select two combinations of the hypergraph model and weighting scheme for classification and clustering. Zhou's normalized hypergraph with VOLUME and TRACE are used for classification. Clique expansion with TRACE and Zhou's normalized hypegraph with CENTROID are used for clustering. The results reported in Tables \ref{C_classify} and \ref{C_cluster} demonstrate that such simple combination can get very promising performance in comparison with the state-of-the-art algorithms. Clearly, such phenomena not only verify the importance of the hyperedge weight in hypergraph learning, but also show the potential of hypegraph learning for addressing the visual tasks.
\end{enumerate}

\section{Conclusion} \label{s6}
We presented a comprehensive experimental study of hyperedge weight in hypergraph learning to draw the researchers' attention to the importance of designing hyperedge weights. In order to verify the importance of hyperedge weight, three novel hyperedge weights, namely VOLUME, TRACE and LLRE, are respectively proposed from the perspectives of geometry, multivariate statistical analysis and linear regression. These three novel hyperedge weights and three other commonly adopted hyperedge weights are applied to three popular hypegraph frameworks, including Zhou's normalized Laplacian, clique expansion and star expansion, for two fundamental learning tasks: clustering and classification. Extensive experiments on ORL, COIL20, JAFFE, Sheffield databases demonstrated that a good hyperedge weight can significantly improve the performances of hypergraph learning. Moreover, we compare the simple combination of a conventional hypergraph framework and a carefully chosen weight scheme with some state-of-the-art algorithms in image classification and clustering on two more larger and challenging databases, namely Scene15 and Caltech256. The results show that such simple combination can get a promising performance. Our work is a fundamental study, so there are many meaningful works can be done based on our study. Applying the hypergraph frameworks and new hyperedge weights to address dimensionality reduction~\cite{dhlp}, feature selection~\cite{nmi} and multi-label classification~\cite{mhyper} may be our future works.

\section*{Acknowledgement}
This work has been partially funded by Fundamental Research Funds for the Central Universities (Grant No. CDJXS11181162) and the National Natural Science Foundation of China (Grant No. 91118005). The authors would like to thank the reviewers for their useful comments.




\bibliographystyle{elsarticle-num}
\bibliography{egbib}

\begin{thebibliography}{10}
\expandafter\ifx\csname url\endcsname\relax
  \def\url#1{\texttt{#1}}\fi
\expandafter\ifx\csname urlprefix\endcsname\relax\def\urlprefix{URL }\fi
\expandafter\ifx\csname href\endcsname\relax
  \def\href#1#2{#2} \def\path#1{#1}\fi

\bibitem{lsc}
S.~Gao, I.~Tsang, L.~Chia, Laplacian sparse coding, hypergraph laplacian sparse
  coding, and applications, IEEE Transactions on Pattern Analysis and Machine
  Intelligence 35~(1) (2013) 92--104.

\bibitem{sum}
Y.~Gao, M.~Wang, D.~Tao, R.~Ji, Q.~Dai, 3-d object retrieval and recognition
  with hypergraph analysis, IEEE Transactions on Image Processing 21~(9) (2012)
  4290--4303.

\bibitem{phr}
Y.~Huang, Q.~Liu, S.~Zhang, D.~N. Metaxas, Image retrieval via probabilistic
  hypergraph ranking, in: IEEE conference on Computer Vision and Pattern
  Recognition (CVPR), 2010, pp. 3376--3383.

\bibitem{higher}
P.~Ochs, T.~Brox, Higher order motion models and spectral clustering, in: IEEE
  conference on Computer Vision and Pattern Recognition (CVPR), 2012, pp.
  614--621.

\bibitem{supervised}
T.~Parag, A.~Elgammal, Supervised hypergraph labeling, in: IEEE conference on
  Computer Vision and Pattern Recognition (CVPR), 2011, pp. 2289--2296.

\bibitem{adaptive}
J.~Yu, D.~Tao, M.~Wang, Adaptive hypergraph learning and its application in
  image classification, IEEE Transactions on Image Processing 21~(7) (2012)
  3262--3272.

\bibitem{nmi}
Z.~Zhang, P.~Ren, E.~Hancock, Unsupervised feature selection via hypergraph
  embedding, in: Birtish Machine Vision Conference (BMVC), 2012, pp. 1--11.

\bibitem{he}
L.~Pu, B.~Faltings, Hypergraph learning with hyperedge expansion, in: Machine
  Learning and Knowledge Discovery in Databases, 2012, pp. 410--425.

\bibitem{expansion}
J.~Y. Zien, M.~D. Schlag, P.~K. Chan, Multilevel spectral hypergraph
  partitioning with arbitrary vertex sizes, IEEE Transactions on Computer-Aided
  Design of Integrated Circuits and Systems 18~(9) (1999) 1389--1399.

\bibitem{mean}
S.~Agarwal, J.~Lim, L.~Zelnik-Manor, P.~Perona, D.~Kriegman, S.~Belongie,
  Beyond pairwise clustering, in: IEEE conference on Computer Vision and
  Pattern Recognition (CVPR), Vol.~2, 2005, pp. 838--845.

\bibitem{zhou}
D.~Zhou, J.~Huang, B.~Sch{\"o}lkopf, Learning with hypergraphs: Clustering,
  classification, and embedding, in: Advances in neural information processing
  systems (NIPS), 2006, pp. 1601--1608.

\bibitem{bolla}
M.~Bolla, Spectra, euclidean representations and clusterings of hypergraphs,
  Discrete Mathematics 117.

\bibitem{hol}
S.~Agarwal, K.~Branson, S.~Belongie, Higher order learning with graphs, in:
  International Conference on Machine Learning (ICML), 2006, pp. 17--24.

\bibitem{rodriguez}
J.~Rodr{\'\i}guez, On the laplacian spectrum and walk-regular hypergraphs,
  Linear and Multilinear Algebra 51.

\bibitem{lpi}
D.~Cai, X.~He, J.~Han, Document clustering using locality preserving indexing,
  IEEE Transactions on Knowledge and Data Engineering 17~(12) (2005)
  1624--1637.

\bibitem{gnmf}
D.~Cai, X.~He, J.~Han, T.~S. Huang, Graph regularized nonnegative matrix
  factorization for data representation, IEEE Transactions on Pattern Analysis
  and Machine Intelligence 33~(8) (2011) 1548--1560.

\bibitem{lpp}
X.~He, S.~Yan, Y.~Hu, P.~Niyogi, H.-J. Zhang, Face recognition using
  laplacianfaces, IEEE Transactions on Pattern Analysis and Machine
  Intelligence 27~(3) (2005) 328--340.

\bibitem{glpp}
S.~Huang, A.~Elgammal, L.~Huangfu, D.~Yang, X.~Zhang, Globality-locality
  preserving projections for biometric data dimensionality reduction, in: IEEE
  Conference on Computer Vision and Pattern Recognition Workshops (CVPRW),
  2014, pp. 15--20.

\bibitem{sparse}
J.~Wright, A.~Y. Yang, A.~Ganesh, S.~S. Sastry, Y.~Ma, Robust face recognition
  via sparse representation, IEEE Transactions on Pattern Analysis and Machine
  Intelligence 31~(2) (2009) 210--227.

\bibitem{collabrative}
L.~Zhang, M.~Yang, X.~Feng, Sparse representation or collaborative
  representation: Which helps face recognition?, in: International Conference
  on Computer Vision (ICCV), 2011, pp. 471--478.

\bibitem{det}
P.~Gritzmann, V.~Klee, On the complexity of some basic problems in
  computational convexity, in: Polytopes: Abstract, Convex and Computational,
  1994, pp. 373--466.

\bibitem{sgraph}
R.~Timofte, L.~Van~Gool, Sparse representation based projections, in: Birtish
  Machine Vision Conference (BMVC), 2011, pp. 61--1.

\bibitem{orl}
F.~S. Samaria, A.~C. Harter, Parameterisation of a stochastic model for human
  face identification, in: IEEE Workshop on Applications of Computer Vision,
  1994.

\bibitem{coil20}
S.~A. Nene, S.~K. Nayar, H.~Murase, Columbia object image library (coil-20),
  Technical Report CUCS-005-96.

\bibitem{umist}
D.~B. Graham, N.~M. Allinson, Face recognition: From theory to applications,
  NATO ASI Series F, Computer and Systems Sciences 163.

\bibitem{jaffe}
M.~N. Dailey, C.~Joyce, M.~J. Lyons, M.~Kamachi, H.~Ishi, J.~Gyoba, G.~W.
  Cottrell, Evidence and a computational explanation of cultural differences in
  facial expression recognition., Emotion 10.

\bibitem{sence15}
S.~Lazebnik, C.~Schmid, J.~Ponce, Beyond bags of features: Spatial pyramid
  matching for recognizing natural scene categories, in: IEEE conference on
  Computer Vision and Pattern Recognition (CVPR), Vol.~2, 2006, pp. 2169--2178.

\bibitem{caltech256}
G.~Griffin, A.~Holub, P.~Perona, Caltech-256 object category dataset.

\bibitem{picodes}
A.~Bergamo, L.~Torresani, A.~Fitzgibbon, Picodes: Learning a compact code for
  novel-category recognition, in: Advances in neural information processing
  systems (NIPS), 2011, pp. 2088--2096.

\bibitem{rcr}
M.~Yang, D.~Zhang, S.~Wang, Relaxed collaborative representation for pattern
  classification, in: IEEE conference on Computer Vision and Pattern
  Recognition (CVPR), 2012, pp. 2224--2231.

\bibitem{rf}
L.~Breiman, Random forests, Machine learning 45~(1) (2001) 5--32.

\bibitem{lbsc}
X.~Chen, D.~Cai, Large scale spectral clustering with landmark-based
  representation., in: AAAI Conference on Artificial Intelligence (AAAI), 2011.

\bibitem{ncut}
J.~Shi, J.~Malik, Normalized cuts and image segmentation, IEEE Transactions on
  Pattern Analysis and Machine Intelligence 22~(8) (2000) 888--905.

\bibitem{eigenmap}
M.~Belkin, P.~Niyogi, Laplacian eigenmaps for dimensionality reduction and data
  representation, Neural computation 15~(6) (2003) 1373--1396.

\bibitem{dhlp}
S.~Huang, D.~Yang, Y.~Ge, D.~Zhao, X.~Feng, Discriminant hyper-laplacian
  projections with its applications to face recognition, in: IEEE conference on
  Multimedia and Expo Workshop on HIM (ICMEW), 2014.

\bibitem{mhyper}
L.~Sun, S.~Ji, J.~Ye, Hypergraph spectral learning for multi-label
  classification, in: ACM international conference on Knowledge discovery and
  data mining (SIGKDD), 2008, pp. 668--676.

\end{thebibliography}






\end{document}